\documentclass[a4paper,11pt]{article}
\usepackage[utf8]{inputenc}
\usepackage[T1]{fontenc}
\usepackage[margin=1in]{geometry}

\usepackage[colorlinks,linkcolor=blue,citecolor=blue]{hyperref}
\usepackage{amsfonts,amsmath,amsthm,amssymb}
\usepackage{mathtools}

\usepackage{graphicx}
\usepackage{booktabs} %
\usepackage{multirow} %
\usepackage[noblocks]{authblk}
\usepackage{enumitem}
\usepackage{bm}
\usepackage{xspace}
\usepackage{nicefrac}
\usepackage{algorithm}
\usepackage[noend]{algpseudocode}

\usepackage[numbers]{natbib}
\usepackage{array}
\newcommand{\head}[2]{\multicolumn{1}{>{\centering\arraybackslash}p{#1}}{\textsc{#2}}}

\usepackage{amsmath}
\usepackage{amsthm} %
\usepackage{mathtools,thmtools}
\usepackage{amsfonts}

\usepackage{enumitem}
\usepackage{bm}
\usepackage{xspace}
\usepackage{nicefrac}
\usepackage[capitalise]{cleveref}
\usepackage{xcolor}
\usepackage{dsfont}

\usepackage[textsize=footnotesize,textwidth=0.75in,
            color=cyan,bordercolor=white]{todonotes}
\presetkeys{todonotes}{size=\footnotesize\sffamily\bfseries\boldmath\color{white}}{}

\newcommand{\ignore}[1]{}

\declaretheoremstyle[
	    spaceabove=\topsep,
	    spacebelow=\topsep,
	    bodyfont=\normalfont\itshape,
    ]{theorem}

\declaretheorem[style=theorem,name=Theorem]{theorem}

\declaretheoremstyle[
	    spaceabove=\topsep,
	    spacebelow=\topsep,
	    bodyfont=\normalfont,
    ]{definition}

\declaretheorem[style=theorem,sibling=theorem,name=Claim]{claim}
\declaretheorem[style=theorem,sibling=theorem,name=Proposition]{proposition}

\declaretheorem[style=theorem,numbered=no,name=Theorem]{theorem*}
\declaretheorem[style=theorem,numbered=no,name=Lemma]{lemma*}
\declaretheorem[style=theorem,numbered=no,name=Corollary]{corollary*}
\declaretheorem[style=theorem,numbered=no,name=Proposition]{proposition*}
\declaretheorem[style=theorem,numbered=no,name=Claim]{claim*}
\declaretheorem[style=theorem,numbered=no,name=Fact]{fact*}
\declaretheorem[style=theorem,numbered=no,name=Observation]{observation*}
\declaretheorem[style=theorem,numbered=no,name=Conjecture]{conjecture*}

\declaretheorem[style=definition,numbered=no,name=Definition]{definition*}
\declaretheorem[style=definition,numbered=no,name=Remark]{remark*}
\declaretheorem[style=definition,numbered=no,name=Example]{example*}
\declaretheorem[style=definition,numbered=no,name=Question]{question*}

\DeclareMathAlphabet{\mathbfsf}{\encodingdefault}{\sfdefault}{bx}{n}

\DeclareMathOperator*{\trace}{Tr}
\DeclareMathOperator*{\diag}{diag}
\newcommand{\lr}[1]{\mathopen{}\left(#1\right)}
\newcommand{\Lr}[1]{\mathopen{}\big(#1\big)}
\newcommand{\LR}[1]{\mathopen{}\Big(#1\Big)}

\newcommand{\norm}[1]{\|#1\|}

\newcommand{\set}[1]{\{#1\}}

\newcommand{\Lrset}[1]{\mathopen{}\big\{#1\big\}}

\newcommand{\abs}[1]{|#1|}

\newcommand{\floor}[1]{\lfloor #1 \rfloor}

\renewcommand{\O}{O}

\newcommand{\E}{\mathbb{E}}

\newcommand{\tr}{^{\mkern-1.5mu\scriptstyle\mathsf{T}}}

\newcommand{\st}{\star}

\newcommand{\reals}{\mathbb{R}}

\let\oldtfrac\tfrac
\renewcommand{\tfrac}[2]{\smash{\oldtfrac{#1}{#2}}}

\let\nablaold\nabla
\renewcommand{\nabla}{\nablaold\mkern-1mu}

\floatname{algorithm}{\NAME\kern-0.7ex-\kern-0.7ex}

\newcommand*\samethanks[1][\value{footnote}]{\footnotemark[#1]}

\title{\mbox{Memory-Efficient Adaptive Optimization}}

\author{%
Rohan Anil%
\thanks{Google Brain, Mountain View. Emails: \texttt{\{rohananil,vineet,tkoren\}@google\!.\!com}.}
\quad
Vineet Gupta%
\samethanks[1]
\quad
Tomer Koren%
\samethanks[1]
\quad
Yoram Singer%
\thanks{Princeton University. Email: \texttt{y.s@cs.princeton\!.\!edu}.}
}

\def\NAME{SM3\xspace}
\crefname{sms}{SM3}{SM3s}
\crefformat{sms}{SM3-#2#1#3}

\floatname{algorithm}{\NAME\!-\kern-0.5ex}

\algnewcommand{\LineComment}[1]{\State \(\triangleright\) #1}

\crefname{figure}{Figure}{Figures}

\newcommand{\mx}[1]{\mu_{#1}}
\newcommand{\mi}[1]{\nu_{#1}}
\newcommand{\tmx}[1]{\mu'_{#1}}
\newcommand{\tmi}[1]{\nu'_{#1}}
\newcommand{\onev}[1]{\bm{1}_{#1}}
\newcommand{\entode}{en$\to$de\xspace}
\newcommand{\entofr}{en$\to$fr\xspace}

\begin{document}
\maketitle

\begin{abstract}
Adaptive gradient-based optimizers such as Adagrad and Adam are crucial for
achieving state-of-the-art performance in machine translation and language
modeling. However, these methods maintain second-order statistics for each
parameter, thus introducing significant memory overheads that restrict the
size of the model being used as well as the number of examples in a
mini-batch. We describe an effective and flexible adaptive optimization method
with greatly reduced memory overhead. Our method retains the benefits of
per-parameter adaptivity while allowing significantly larger models and batch
sizes. We give convergence guarantees for our method, and demonstrate its
effectiveness in training very large translation and language models with up
to 2-fold speedups compared to the state-of-the-art.
\end{abstract}

\section{Introduction}

Adaptive gradient-based optimizers such as Adagrad \cite{duchi2011adaptive} and
Adam \cite{kingma2014adam} are among the de facto methods of choice in modern
machine learning. These methods adaptively tune the learning rate for each
parameter during the optimization process using cumulative second-order
statistics. Often offering superior convergence properties, these methods are
very attractive in large scale applications due to their moderate time and space
requirements, which are linear in the number of parameters.
However, when training extremely large models even the modest memory overhead
imposes grave limitations on the quality of the trained model. For example,
recent advances in natural language processing \cite{vaswani2017attention, gpt2}
show that models with hundreds of millions to billions of parameters, trained
with adaptive optimization methods, achieve state-of-the-art results. In such
cases, the memory overhead of the optimizer severely restricts the size of the
model that can be used as well as the number of examples in each mini-batch,
both of which have a dramatic effect on the accuracy of the model.

Motivated by these challenges, we describe an adaptive optimization method that
retains the benefits of standard per-parameter adaptivity while significantly
reducing memory overhead. Our construction is general and flexible, and very
simple to implement. We give convergence guarantees for our method in the convex
(online or stochastic) optimization setting, and demonstrate experimentally that
it is particularly effective when the gradients exhibit natural {\em activation
patterns}; namely, when the parameters can be subdivided into (not necessarily
disjoint) sets where gradient entries within sets are correlated and of a
similar order of magnitude. For example, we often observe in deep networks that
the incoming (outgoing) edges into (from) a neuron are jointly activated and,
loosely speaking, their associated gradients exhibit similar statistical
characteristics. That said, the analysis of our optimization algorithm makes no
statistical assumptions on the gradients and is applicable in general stochastic
convex optimization settings. Further, we do not assume that the activation
pattern is fully prescribed a-priori to the algorithm.

Large scale experiments show that our algorithm achieves comparable, and at
times superior, rates of convergence compared to standard linear-space
adaptive methods. Focusing primarily on language modeling tasks where
state-of-the-art models are extremely large, we further demonstrate that the
reduction in memory footprint can be utilized for a substantial increase in
the batch size, which greatly speeds up convergence in a distributed
environment. For a fixed budget of computational resource our method is able
to shorten the end to end wall-time for convergence by up to 50\%. Our method
exhibits slightly improved per-step time. The latter could be attributed to
reduction in the frequency of memory accesses.

\subsection{Related work}

Adaptive learning rates in online and stochastic optimization date back at least
to~\cite{auer2002adaptive} and were popularized in
\cite{duchi2011adaptive,mcmahan2010adaptive}, the former of which introduced the
well-known Adagrad algorithm. Several variants of Adagrad have now been proposed
in the optimization and machine learning literature (see
\citep{reddi2018convergence} and the references therein), the most notable of
which is Adam~\cite{kingma2014adam}. All of these methods require (at least)
linear space for maintaining various per-parameter statistics during their
execution.
One notable exception, which is directly related to our work, is the Adafactor
algorithm~\cite{shazeer18} that was proposed as a way to reduce the memory costs
of Adam, primarily for training large language models. While the memory
requirements of our construction are similar to Adafactor's, the application
scope and the convergence properties of the two algorithms are quite different.
We discuss the relationship in more detail in~\cref{sec:patterns} and give an
empirical comparison between the algorithms in~\cref{sec:exper}.

Another closely related method is the Shampoo~\cite{shampoo-icml} algorithm for
optimization over tensor structures. Seemingly, the goal of Shampoo is very
different from ours: going beyond entry-wise learning rates and employing
\emph{full-matrix} regularization in a computationally efficient way.
Nonetheless, Shampoo can also be seen as a method to substantially reduce the
memory footprint of full-matrix preconditioned algorithms (specifically,
full-matrix Adagrad). In a sense, our algorithms are analogous to a diagonalized
version of the Shampoo algorithm.
Yet another recent adaptive optimization method is the GGT
algorithm~\citep{GGT}. Similarly to Shampoo, the goal of the latter is to reduce
the computation cost of full-matrix preconditioning in order to make it
practical in large scale settings. However, GGT stores multiple copies of the
gradient over the course of its execution, and as a result, its space
requirements restricts it from being applied at large scale.

\section{Preliminaries}
\label{sec:prelim}

\subsection{Online optimization}
We henceforth assume the general {\em online optimization} setting (see
\citep{shalev2012online,hazan2016introduction}). Online optimization consists of
rounds $t=1,\ldots,T$, where in each round the algorithm chooses a parameter
vector $w_t \in \reals^d$. After making a choice on round $t$, the algorithm
receives a loss function $\ell_t : \reals^d \to \reals$ which is used to form an
update of the parameters. In our analysis, we focus on online {\em convex}
optimization in which $\ell_1,\ldots,\ell_T$ are convex. Often, as is the case
in this paper, the update is determined by the gradient $g_t=\nabla\ell_t(w_t)$
of the instantaneous loss $\ell_t$ at the current iterate $w_t$. The algorithm
is measured by its $T$-round regret with respect to a given comparator $w^\star
\in \reals^d$, defined as the quantity
$
  \sum_{t=1}^T \ell_t(w_t) - \sum_{t=1}^T \ell_t(w^\star)
  .
$
An online optimization algorithm is convergent if its regret is $o(T)$, i.e.,
its average regret approaches zero as $T$ grows.

The above setting includes stochastic (possibly mini-batched) optimization as a
special case. In stochastic optimization the underlying goal is to minimize a
population loss $L(w) = \E_{z \sim D} [\ell(w,z)]$ based on samples of $z$. Here
$\ell(w,z)$ defines the loss of parameters $w$ w.r.t a batch $z$. The online
loss function $\ell_t(w) = \ell(w,z_t)$ is the average loss over a mini-batch
$z_t$ received on iteration $t$. The stochastic gradient $g_t$ is a
conditionally unbiased estimate of the gradient of $L$ at the current parameter
vector $w_t$. Under convexity assumptions, an online algorithm with vanishing
average regret can be converted to a stochastic optimization algorithm for
minimizing the population loss~$L$~\citep{cesa2004generalization}.

\subsection{Adaptive methods}
For the sake of self-containment, we give a brief description of adaptive
gradient methods, focusing on Adagrad~\cite{duchi2011adaptive}. Adagrad
maintains at each step $t$ parameter-wise accumulated statistics which are
computed from the previously obtained gradients $g_1,\ldots,g_t$:
\begin{align} \label{eq:accum}
  \gamma_t(i)
  =
  \sum_{s=1}^t g_s^2(i)
  ~,\qquad
  \forall ~ i \in [d]
  ~.
\end{align}
Based on these statistics, the update rule of the algorithm on step $t$ takes
the form:
\begin{align*}
  w_{t+1}(i)
  =
  w_{t}(i) - \eta \frac{g_t(i)}{\sqrt{\gamma_t(i)}}
  ~,\qquad
  \forall ~ i \in [d]
  ~,
\end{align*}
where $\eta>0$ is an external learning rate parameter.
\citet{duchi2011adaptive} proved the following regret bound for Adagrad with
respect to a given $w^\star$ (with properly tuned $\eta$):
\begin{align} \label{eq:regret-adagrad}
  \sum_{t=1}^T \ell_t(w_t) - \sum_{t=1}^T \ell_t(w^\star)
  =
  \O\lr{ D \sum_{i=1}^d \sqrt{\sum_{t=1}^T g_t^2(j)} }
  ,
\end{align}
where $D \geq \max_t \norm{w_t-w^\st}_\infty$. Adagrad has proved to be
particularly useful in training sparse models, where the effective learning
rates $\eta\big/\!\sqrt{\gamma_t(i)}$ decay in a moderate way for rare, yet
potentially informative, features. In these settings, Adagrad can potentially
lead to substantial improvements in convergence time; see for instance the
discussion in~\cite{duchi2011adaptive}. Crucially, however, Adagrad must
maintain auxiliary sequence of accumulators $\gamma_t$ and thus needs
$\Omega(d)$ additional space. The goal of this paper is to provide
memory-efficient methods with comparable convergence characteristics that
refrain from maintaining the full vectors $\gamma_t$.

\section{The \NAME Algorithm}
\label{sec:algo}

We now present our memory-efficient adaptive optimization algorithm. As an
abstraction, the algorithm employs a \emph{cover} of the parameters: a
collection of $k$ nonempty sets $\set{S_r}_{r=1}^k$, such that
$S_r\subseteq[d]$ and $\cup_r S_r = [d]$. In particular, each index $i\in[d]$
may be contained in multiple sets $S_r$. The algorithm maintains a single
variable for each set $S_r$ in the cover. Thus, the additional space it
requires is $O(k)$ rather than the $O(d)$ required by standard adaptive
methods. In large scale applications, $k$ will be chosen to be negligible in
comparison to $d$, which would translates to substantial savings in memory;
see~\cref{sec:patterns} for a discussion on the covers used in practice.

Concretely, for each set $S_r$ in the cover, the algorithm maintains a running
sum, $\mx{t}(r)$, of the \emph{maximal} variance over all gradient entries $j
\in S_r$. Next, for each parameter $i$, we take the \emph{minimum} over all
variables $\mx{t}(r)$ associated with sets which cover $i$, denoted $S_r \ni i$.
Thereafter, the learning rate corresponding to the $i$'th gradient entry is
determined by taking the square-root of this minimum, denoted by $\mi{t}(i)$.
Accordingly, we name our algorithm the \emph{S}quare-root of \emph{M}inima of
\emph{S}ums of \emph{M}axima of \emph{S}quared-gradients \emph{M}ethod, or in
short, \emph{\NAME}. See Algorithm~\cref{alg:alg} for its pseudocode.

\begin{algorithm}[H]
\begin{algorithmic}[1]
  \State {\bf parameters:} learning rate $\eta$
  \State initialize $w_1 = 0$ ;~ $\forall r \in [k]: \, \mx{0}(r) = 0$
  \For{$t = 1,\ldots,T$}
    \State receive gradient $g_t = \nabla \ell_t(w_t)$
    \For {$r = 1,\ldots,k$}
      \State set {$\mx{t}(r) \gets \mx{t-1}(r) + \max_{j \in S_r} g^2_{t}(j)$}
    \EndFor
    \For{$i = 1,\ldots,d$}
      \State set $\mi{t}(i) \gets \min_{r : S_r \ni i} \mx{t}(r)$
      \State update $w_{t+1}(i) \gets w_{t}(i) - \eta \, g_{t}(i) \big/\! \sqrt{\mi{t}(i)}$
      \Comment{with the convention that $0/0 = 0$}
    \EndFor
  \EndFor
\end{algorithmic}
\caption{}
\label[sms]{alg:alg}
\end{algorithm}

As noted above, \cref{alg:alg} requires only $O(k)$ space in addition to the
space required for storing the parameters $w_t$ themselves. The time per
iteration of \cref{alg:alg} is $O(\sum_{r=1}^k \abs{S_r})$. To see this,
consider a bipartite graph defined over $d+k$ vertices. Nodes on one side of the
graph correspond to indices $i \in [d]$, while nodes on the other side
correspond to indices $j\in[k]$. The edges of the graphs are all pairs $(i,j)$
such that $i \in S_j$.  The complexity of each inner for-loop of the
algorithm scales with the number of edges in this graph, which is equal to
$O(\smash{\sum_{r=1}^k} \abs{S_r})$. Note that updating the weights $w_t$ takes
$O(d)$ time, which is always dominated by the former quantity.

The following provides convergence guarantees for \cref{alg:alg}.
\begin{proposition} \label{thm:main}
Assume that the loss functions $\ell_1,\ell_2,\ldots$ are convex, and let
$w_1,w_2,\ldots$ be the iterates generated by \cref{alg:alg}. Then, for any
$w^\st \in \reals^d$,
\begin{align*}
  \sum_{t=1}^T \Lr{ \ell_t(w_t) - \ell_t(w^\star) }
  \leq
  2 D \sum_{i=1}^d \sqrt{\min_{r : S_r \ni i} \sum_{t=1}^T \max_{j \in S_r} g_t^2(j)}
  ~,
\end{align*}
where $\max_{t} \norm{w_t-w^\st}_\infty \leq D$ and choosing $\eta
= D$.%
\footnote{Here we implicitly assume that the iterates of \cref{alg:alg} remain
bounded and $D$ is a constant. This can be enforced by projecting the iterates
to a bounded set of choice; we avoid introducing projections explicitly as they
are rarely used in practice.}
\end{proposition}

For stochastic optimization, i.e., when the functions $\ell_t$ correspond to
i.i.d.~samples with $\E[\ell_t(w)] = L(w)$, the above bound translates via
standard arguments to a $O(1/\sqrt{T})$-type convergence guarantee for the
average iterate $\overline{w}_T = \frac{1}{T} \sum_{t=1}^T w_t$ of the form
\begin{align*}
  \E[L(\overline{w}_T)] - L(w^\star)
  =
  \O\lr{
  \frac{1}{T} \sum_{i=1}^d \E \sqrt{\min_{r : S_r \ni i} \sum_{t=1}^T \max_{j \in S_r} g_t^2(j)} }
  \!.
\end{align*}

Note that adding more sets $S_r$ to the cover used by \NAME always improves
its convergence bound, but results in a worse space complexity and a higher
runtime per step.
When $k=d$ and $S_i = \set{i}$ for all~$i \in [d]$, \cref{alg:alg} reduces to
the Adagrad algorithm, and the regret bound in \cref{thm:main} then precisely
recovers the bound attained by Adagrad (recall \cref{eq:regret-adagrad}).
In general, the right-hand side of \cref{thm:main} is never smaller than
Adagrad's regret bound, as expected from a space-restricted scheme (this is a
consequence of \cref{lem:main} below).
Nevertheless, the two bounds can be of similar order of magnitude in practical
scenarios; see \cref{sec:patterns} below for a detailed~discussion.

We now give a proof of \cref{thm:main}. First, we state two elementary
properties of the step sizes the algorithm computes. For a proof, see
\cref{sec:moreproofs}.

\begin{claim} \label{lem:main}
For any $i$, the sequence $\mi{1}(i), \mi{2}(i), \ldots$ is
monotonically increasing, and
$
  \mi{t}(i)
  \geq
  \sum_{s=1}^t g_{s}^2(i)
  .
$
\end{claim}
\vspace{-0.4cm}
\begin{proof}[Proof of \cref{thm:main}]
Let us first assume that $g_1(i) > 0$ for all $i$, so that $\mi{t}(i) > 0$ for
all $i$ and $t \geq 1$ due to \cref{lem:main}.
We start by observing that \cref{alg:alg} performs Online Mirror Descent
updates, where the step on round $t$ uses the positive definite diagonal matrix
$H_t = \diag(\mi{t}^{1/2})$ for regularization. Then, employing a standard
regret bound for the Online Mirror Descent algorithm with time-dependent
regularization (see for instance \citep[Proposition 3]{duchi2011adaptive}), the
regret of the algorithm is bounded by
\begin{align*}
  &\frac{1}{2\eta} \sum_{t=1}^T \Lr{ \norm{w_t-w^\st}_{H_t}^2 - \norm{w_{t+1}-w^\st}_{H_t}^2 }
  + \frac{\eta}{2} \sum_{t=1}^T \Lr{\norm{g_t}_{H_t}^*}^2
  ~.
\end{align*}
Here, $\norm{x}_H = \sqrt{x\tr H x}$ and $\norm{\cdot}^*$ is the corresponding
dual norm, $\norm{x}_H^* = \sqrt{x\tr H^{-1} x}$.
Henceforth, for notational convenience we set $\mi{0}=0$. Simplifying the first
sum above using the fact that $H_t$ are diagonal matrices, we have
\begin{align*}
  \sum_{t=1}^T \Lr{ \norm{w_t-w^\st}_{H_t}^2 - \norm{w_{t+1}-w^\st}_{H_t}^2 }
  &\leq
  \sum_{t=1}^T (\mi{t}^{1/2}-\mi{t-1}^{1/2}) \cdot (w_t-w^\st)^2
  \\
  &\leq
  \sum_{t=1}^T (\mi{t}^{1/2}-\mi{t-1}^{1/2}) \cdot \Lr{\norm{w_t-w^\st}_\infty^2 \onev{d}}
  \\
  &\leq
  \vphantom{\sum^T}
  D^2 \left(\mi{T}^{1/2} \cdot \onev{d}\right)
  \;=\;
  \vphantom{\sum^T}
  D^2 \trace(H_T) ~.
\end{align*}
Now, let $\gamma_t(i) = \sum_{s=1}^t g_s^2(i)$ and consider the positive
definite diagonal matrix $G_t = \diag(\gamma_t^{1/2})$. From \citep[Lemma
2]{shampoo-icml} with $\Phi(G) = \trace(G)$, we have
\begin{align*}
  \sum_{t=1}^T \Lr{\norm{g_t}_{G_t}^*}^2
  \leq
  \sum_{t=1}^T \Lr{\norm{g_t}_{G_T}^*}^2 + \trace(G_T)
  =
  \vphantom{\sum^T}
  \gamma_T^{-1/2} \cdot \gamma_T + \trace(G_T)
  =
  \vphantom{\sum^T}
  2\trace\lr{ G_T }
  .
\end{align*}
Also, from \cref{lem:main} we know that for all $t$, $H_t \succeq G_t$, thus
\begin{align*}
  \sum_{t=1}^T \Lr{\norm{g_t}_{H_t}^*}^2
  \leq
  \sum_{t=1}^T \Lr{\norm{g_t}_{G_t}^*}^2
  \leq
  2\trace(G_T)
  \leq
  2\trace(H_T)
  ~.
\end{align*}
In summary, we have established that
\begin{align*}
  \sum_{t=1}^T \ell_t(w_t) - \ell_t(w^\star)
  \leq
  \lr{\frac{D^2}{2\eta} + \eta} \trace(H_T)
  ~.
\end{align*}
Plugging in $\eta = D$ and the expression for the diagonal elements of $H_T$, we
obtain the claim.

For the degenerate case where the matrices $H_t$ may not be strictly positive
definite, a careful yet technical inspection of the proof above reveals that our
arguments apply to this case as well by replacing inverses with
pseudo-inverses. The rest of the proof remains intact as the algorithm does not
update parameter $i$ on step $t$ if the corresponding diagonal entry in $H_t$ is
zero.
\end{proof}

\subsection{\NAME-II}

We now discuss a slightly more efficient variant of \NAME, which we describe in \cref{alg:alg2}.
It is similar to \cref{alg:alg}, and improves on the latter in the following sense.

\begin{algorithm}[H]
\begin{algorithmic}[1]
  \State {\bf parameters:} learning rate $\eta$
  \State initialize $w_1 = 0$ ;~ $\forall r \in [k]: \, \tmx{0}(r) = 0$
  \For{$t = 1,\ldots,T$}
    \State receive gradient $g_t = \nabla \ell_t(w_t)$
    \State initialize $\tmx{t}(r) = 0$ for all $r \in [k]$
    \For{$i = 1,\ldots,d$}
      \State $\tmi{t}(i) \gets \min_{r : S_r \ni i} \tmx{t-1}(r) + g^2_{t}(i)$
      \State $w_{t+1}(i) \gets w_{t}(i) - \eta \, g_{t}(i) \big/\! \smash{\sqrt{\tmi{t}(i)}}$
      \Comment{with the convention that $0/0 = 0$}
      \For{all $r : S_r \ni i$}
        \State $\tmx{t}(r) \gets \max\set{\tmx{t}(r), \tmi{t}(i)}$
      \EndFor
    \EndFor
  \EndFor
\end{algorithmic}
\caption{}
\label[sms]{alg:alg2}
\end{algorithm}

\begin{proposition} \label{lem:main2}
For any $i \in [d]$, the sequence $\tmi{1}(i), \ldots, \tmi{T}(i)$ is
monotonically increasing. Further, fixing a sequence of gradients
$g_1,\ldots,g_T$, we have for all $t, i$ that
$
  \sum_{s=1}^t g^2_{s}(i)
  \leq
  \tmi{t}(i)
  \leq
  \mi{t}(i)
  ,
$
where $\mi{1}(i), \ldots, \mi{T}(i)$ is the sequence \cref{alg:alg} emits upon
receiving the gradients $g_1,\ldots,g_T$.
\end{proposition}

(See \cref{sec:moreproofs} for a proof.) In other words, \cref{alg:alg2}
provides a tighter upper bound on the cumulative gradient squares than
\cref{alg:alg}. Consequently, we can show, along similar lines to the proof of
\cref{thm:main}, a slightly better bound for \cref{alg:alg2} that scales with
the quantity $\smash{\sum_{i=1}^d \!\sqrt{\tmi{t}(i)}}$, which is always smaller
than the one appearing in the bound of \cref{alg:alg}.

\section{Discussion} \label{sec:patterns}
Thus far, we gave an analysis of \NAME in a worst-case (convex) setting without
placing any further assumptions on the statistical characteristics of the
underlying stochastic gradients. Further, we did not attempt to relate the cover
used by \NAME to properties of the underlying stochastic optimization problem.
It should not come as a surprise that in this general setting, the convergence
of \NAME might be much worse, at least in theory, than its linear-memory
counterpart Adagrad.

\begin{figure}[ht]
\centering
\begin{tabular}{c c c}
\hspace{10pt}
\includegraphics[height=125pt,width=125pt,trim=60 0 0
0]{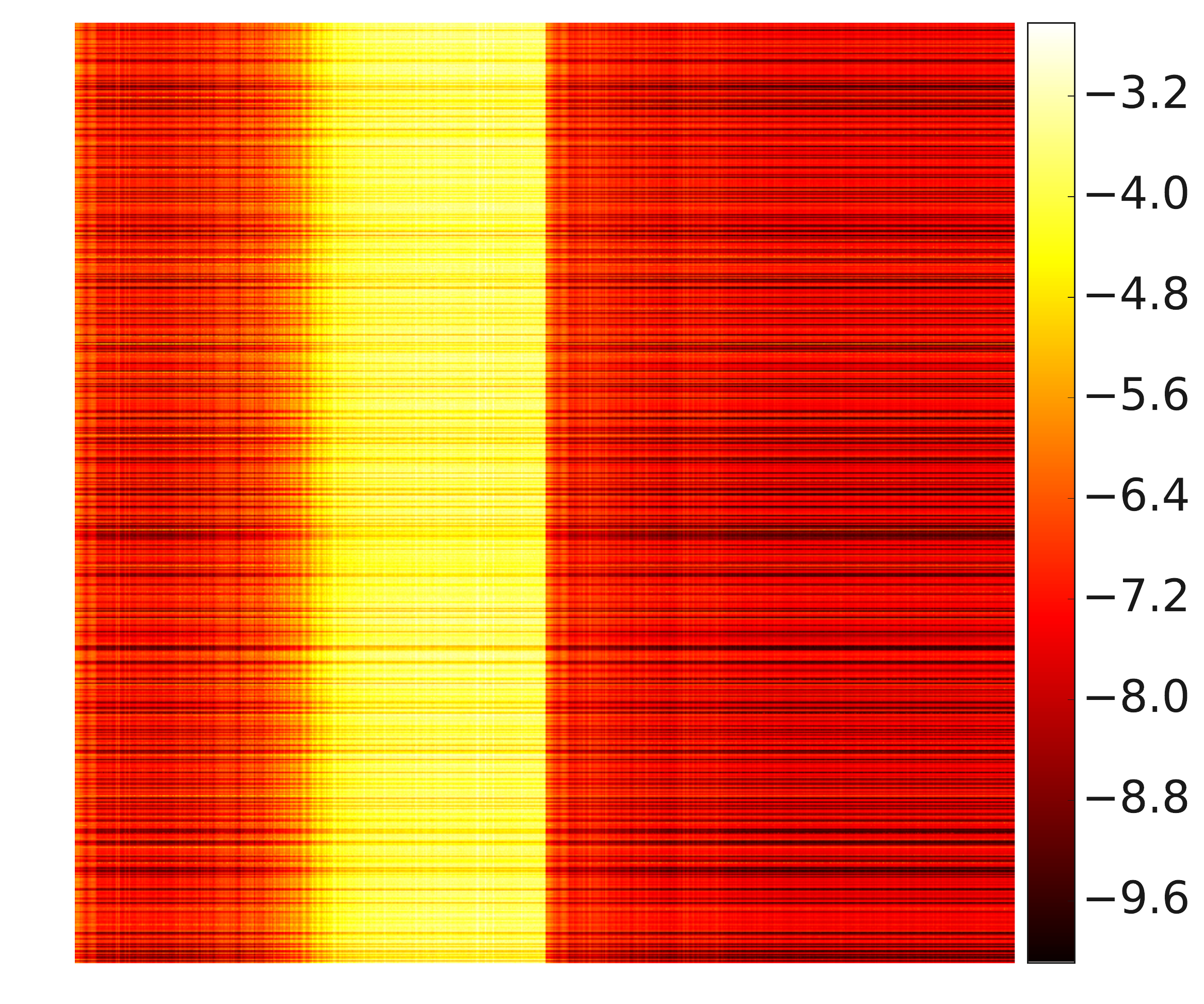} & \hspace{0.02\columnwidth}
\includegraphics[height=125pt,width=125pt,trim=60 0 0
0]{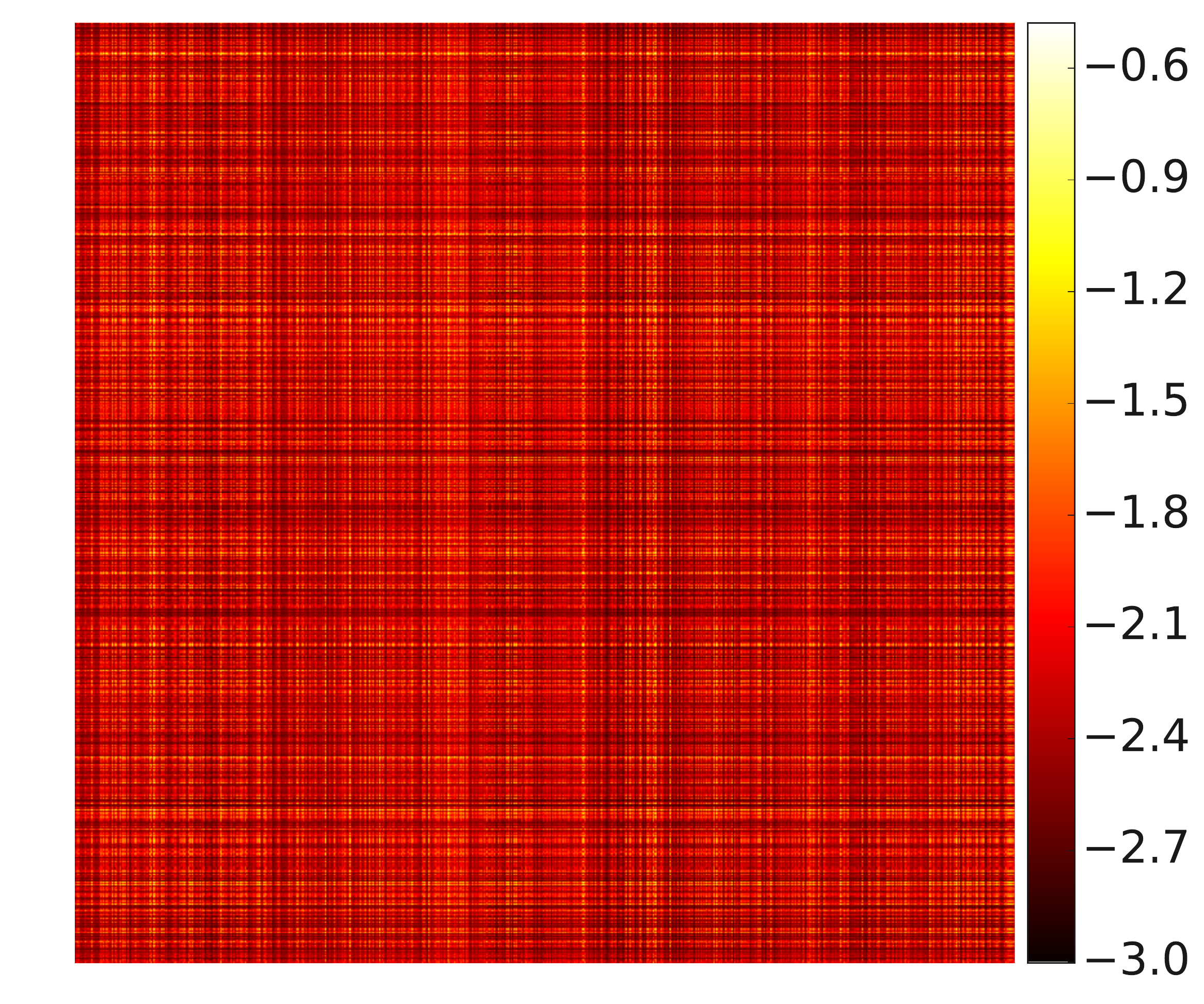} & \hspace{0.02\columnwidth}
\includegraphics[height=125pt,width=125pt,trim=60 0 0
0]{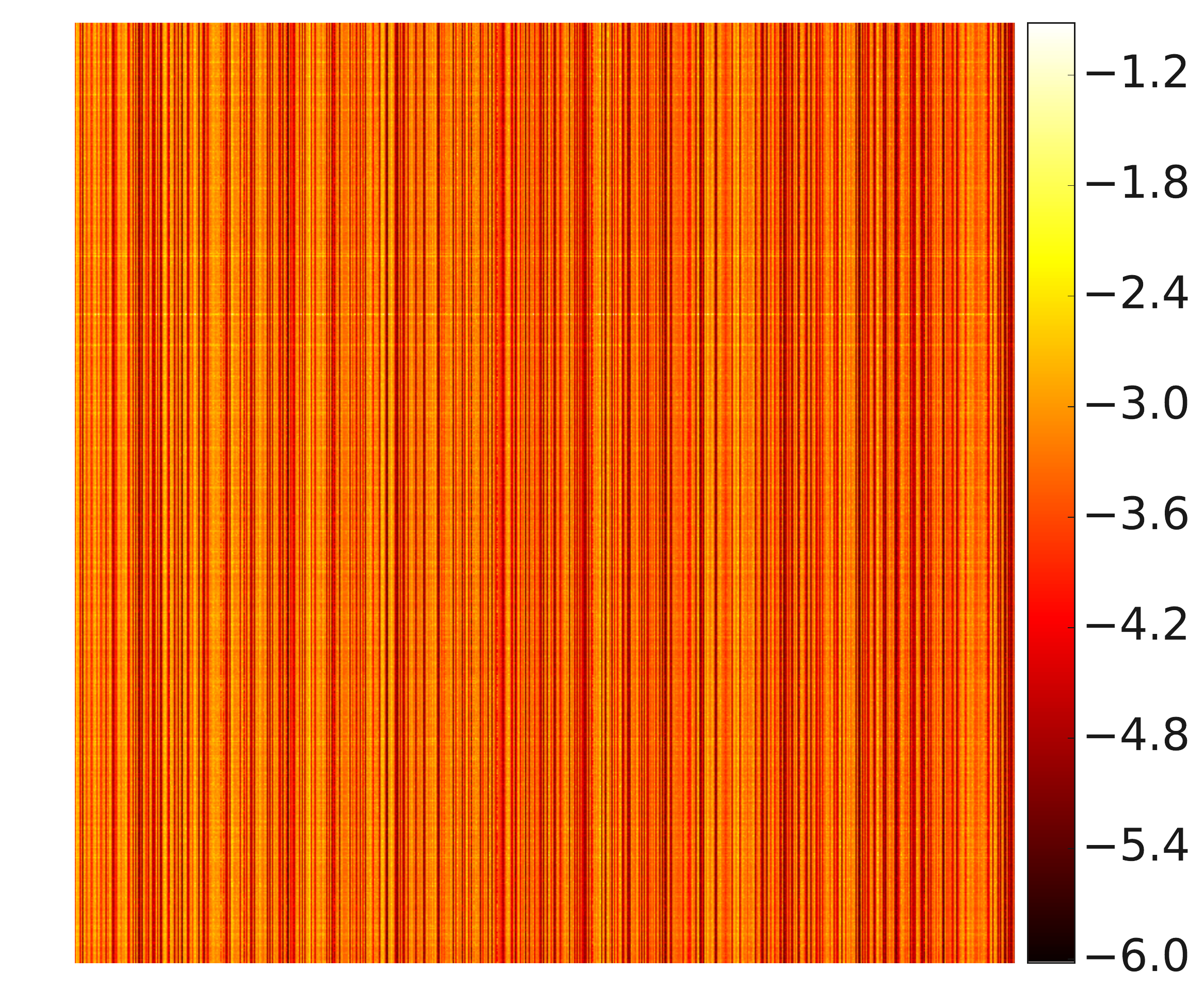} \\
{\hspace{10pt} \small (a) Input embedding \hspace{15pt}}
& {\small (b) Attention layer \hspace{15pt}} & {\small (c) Output softmax
\hspace{15pt}}
\end{tabular}
\caption{Visualization of Adagrad's statistics (cf.~\cref{eq:accum}) for
	different weight matrices in Transformer-Big model trained with Adagrad on
	WMT'14 \entofr (color intensities are in log~scale).}
\label{fig:adagrad_lr}
\end{figure}

\paragraph{Activation patterns.}
Often in our experiments, we observe common statistical attributes that could be
exploited by \NAME. Specifically, we see that certain entries of the stochastic
gradients have (on average) similar values, and exhibit what we refer to as an
{\em activation pattern}. For example, in gradients of embedding layers of deep
networks, an entire row (or column) is either zero or non-zero. Similarly, in
intermediate layers we often observe that gradients associated with the same
unit are of similar order of magnitude. In these cases, a similar phenomenon is
observed in the second-order statistics maintained by adaptive methods.
In \cref{fig:adagrad_lr} we visualize this phenomenon for different layers of a
Transformer network. In \cref{sec:conv-patterns} we give additional
illustrations of similar phenomena in convolutional layers of image
classification models.

\paragraph{Choice of covers.}

The intuitive notion of an activation pattern motivates a natural and generic
choice for the cover used by \NAME in practice. For the parameters of deep
networks, that are organized as a collection of tensors, we form a cover
consisting of slices of co-dimension $1$ for each tensor. Thus, for an $m \times
n$ parameter matrix, the cover consists of rows and columns of the matrix. The
memory requirements therefore drop from $\Theta(mn)$ to merely $\Theta(m+n)$.
For a parameter tensor of dimension $n_1 \times \cdots \times n_p$, the
reduction in memory consumption is even more pronounced, dropping from
$\Theta(\prod_{i=1}^pn_i)$ to $\Theta(\sum_{i=1}^p n_i)$.  This virtually
eliminates the memory overhead associated with maintaining the adaptive learning
rates.

We argue, though only informally, that when choice of cover used by \NAME is
compatible with the observed activation patterns, we expect the convergence of
\NAME to be significantly better, and closely match Adagrad.
Quantitatively, if each parameter $i \in [d]$ is covered by a set $S_r$ such
that $g_s(j) \approx g_s(i)$ for all $j \in S_r$,
then $\max_{j \in S_r} g_{s}^2(j) \approx g_{s}^2(i)$, and thus
$
  \min_{r : S_r \ni i} \sum_{s} \max_{j \in S_r} g_{s}^2(j)
  \approx \sum_{s} g_{s}^2(i)
  ~ .
$
Thus, the bounds in \cref{thm:main} and \cref{eq:regret-adagrad} are of similar
order of magnitude. In other words, in such scenarios we inherit the convergence
properties of Adagrad while using a negligible amount of memory.  We remark that
the activation pattern need not be fully specified in advance; in particular,
\NAME is robust to whether a certain parameter is ``row tied'' or ``column
tied'', as long as both rows and columns are included in the cover.

\paragraph{Comparison with Adafactor.}
Adafactor \cite{shazeer18} is a very effective method for space-efficient
adaptive optimization. \NAME and Adafactor differ in a number of important ways.
First, Adafactor is only defined for matrix-shaped parameters while \NAME
applies to tensors of arbitrary dimensions, and even more generally, to any
predefined cover of the parameters. Second, Adafactor is in essence a fixed
learning-rate algorithm, being a memory-constrained variation of Adam, and often
requires a manually devised learning-rate schedule to ensure convergence. In
contrast, \NAME adapts its learning rates in an adaptive, data-driven manner
similar to Adagrad. Finally, \NAME comes with rigorous convergence guarantees in
stochastic convex optimization settings.

\section{Experiments}
\label{sec:exper}

\def\4x4{4\!\times\!4}
\def\8x8{8\!\times\!8}
\def\16x16{16\!\times\!16}

We demonstrate the practical efficacy of \NAME on several machine learning tasks
using published state-of-the-art architectures. We focus on three domains:
machine translation, language modeling, and image classification.
We implemented \NAME as an optimizer in TensorFlow~\cite{tensorflow}; source code is publicly available at \cite{sm3git2019}. Our
implementation follows the pseudocode of~\cref{alg:alg2}, as it performed
slightly yet consistently better than~\cref{alg:alg} in our experiments (as
predicted by our bounds).
We use covers induced by rows and columns of matrices, and more generally, by
slices of higher-order tensors (e.g., in convolutional layers represented by
$4$-dimensional tensors), as described in~\cref{sec:patterns}. In addition to
being compatible with the natural activation patterns, these covers facilitates
efficient tensor operations available on GPUs and TPUs for computing \emph{max}
and \emph{min} over the sets. In all experiments, we used the Cloud TPU-v2
device~\citep{jouppi2017datacenter} where each core has 8GiB of memory. For more
details on all of our experiments, including the values of hyperparameters used
in each of them, see~\cref{sec:exper-detail}.

\subsection{Machine translation}
We experimented with machine translation tasks on two standard datasets from
WMT'14: English to French (\entofr) with 36.3M sentence pairs, and English to
German (\entode) with 4.5M sentence pairs. We used the state-of-the-art
Transformer architecture~\citet{vaswani2017attention}. The basic version of
this model has 93.3M parameters and consumes 0.36GiB memory. The larger variant
(coined Transformer-Big) has 375.4M parameters (1.432GiB) and consists of 6
layers for its encoder and decoder, where each layer is composed of 1024 model
dimensions, 8192 hidden dimensions, and 16 attention heads.

\begin{figure}[ht]
\begin{center}
\includegraphics[width=0.45\columnwidth]{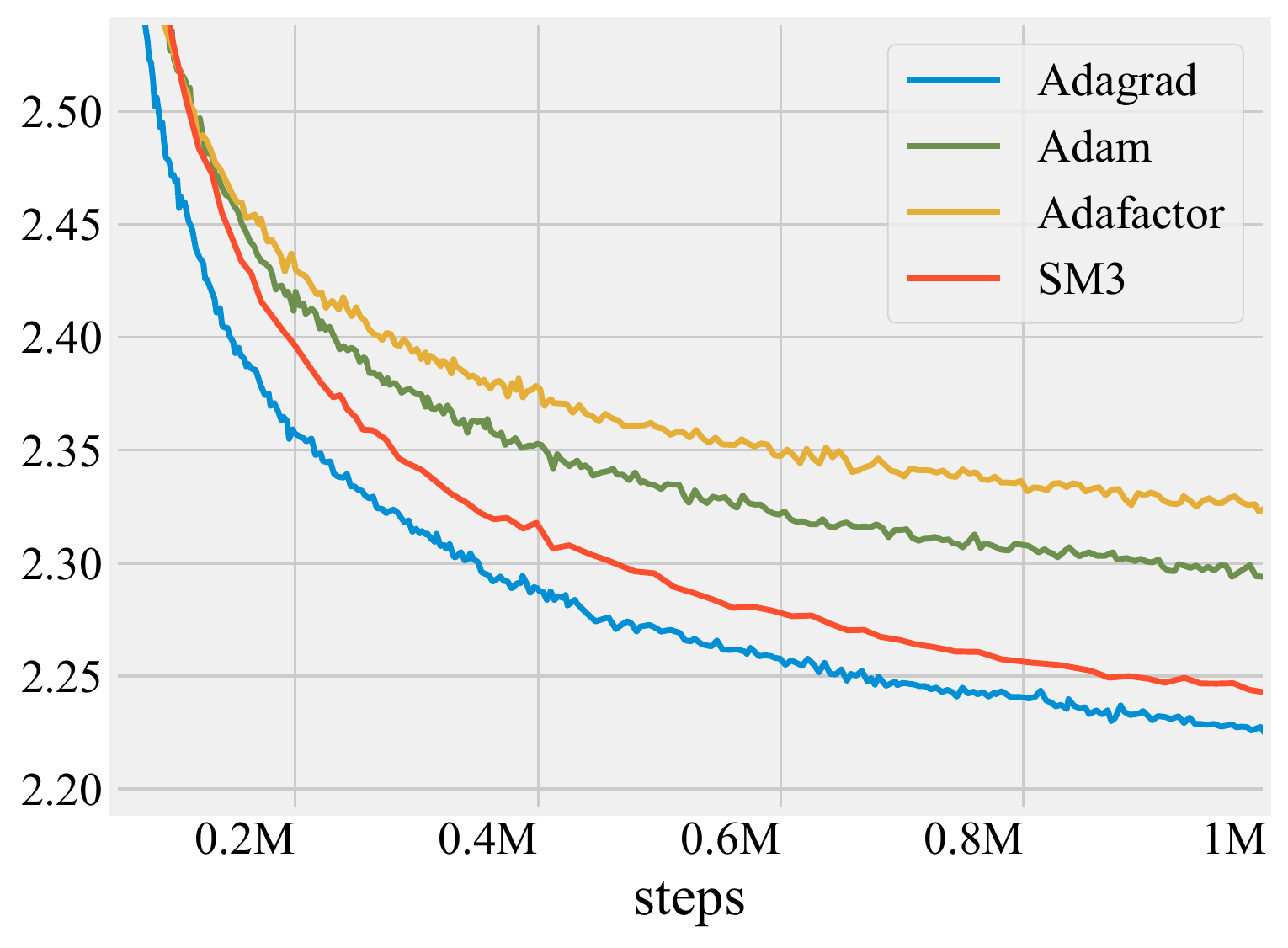}
\hspace{0.05\columnwidth}
\includegraphics[width=0.45\columnwidth]{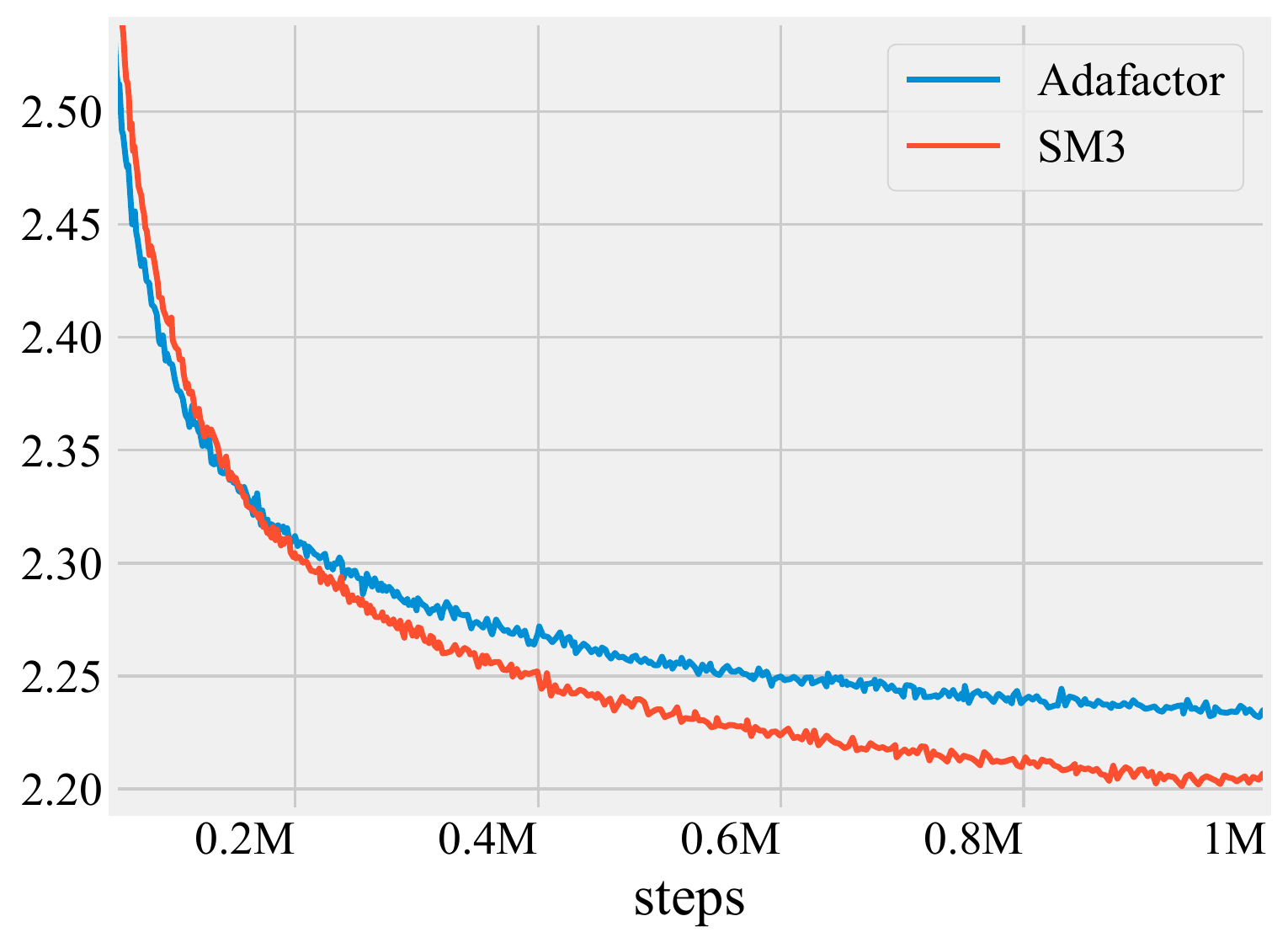}
\end{center}
\vspace{-0.25cm}
\caption{Test log-perplexity of a Transformer-Big model on WMT'14 \entofr, when
training with batch sizes of 384 (left) and 768 (right). For batch size of 768,
Adam and Adagrad were infeasible as they exceeded the available memory.}
\label{en_fr}
\end{figure}

Here we report our results on the larger Transformer-Big, and defer results on
the basic Transformer to~\cref{sec:more-exper}. We trained Transformer-Big on
the \entofr  dataset with batches of size 384, and compared \NAME with several
standard optimizers in each of the tasks. In all cases, we used momentum
(including for Adagrad) and extensively tuned all hyperparameters. We also ran
SGD with momentum (with various exponential decay schedules), but it performed
poorly and hence it is omitted from the figures.
The results are provided in~\cref{en_fr} and~\cref{tbl:bleu}, and demonstrate
that \NAME performed substantially better and provided a large improvement in
BLEU score compared to Adam and Adafactor. In addition, the small memory
requirements of \NAME and Adafactor allowed us to \emph{double the number of
examples in a batch} to a total of 768, with minimal additional computation
resources. In this setting, we found that \NAME outperformed Adafactor in terms
of the number of steps as well as the wall-time to convergence by roughly a
factor of $2$. We further observed that \NAME approximated the Adagrad
second-order statistics tightly. More details are provided in
\cref{sec:more-exper}.

Both models were trained on a $\4x4$ Cloud~TPU-v2 using the
Lingvo~\citep{lingvo} sequence modeling framework, with 32K
word-pieces~\cite{schuster12} for each language pair. BLEU scores were computed
on the Newstest 2014 for evaluation, on tokenized, true-case outputs, and
without manual post-processing of the text, similar to~\cite{wu2016google}. Our
BLEU scores are not directly comparable to those
of~\cite{vaswani2017attention}.  We instead followed the experimental protocol
described in a later work~\cite{chen18}.

\begin{table}[ht]
\begin{center}
\begin{small}
\begin{tabular}{lcccr}
\toprule
\head{2cm}{Optimizer} & \head{3cm}{Batch Size \mbox{per core (total)}} &  \head{2.5cm}{Memory Usage per core} & \head{1.5cm}{BLEU} \\
\midrule
Adam & 12 (384) & 6.88 GiB & $38.96 \pm 0.002$ \\
Adagrad & 12 (384) & 6.85 GiB & $39.90 \pm 0.003$  \\
Adafactor & 12 (384) &  5.43 GiB & $37.89 \pm 0.002$ \\
\NAME & 12 (384) &  5.36 GiB & $39.81	\pm 0.002$  \\
\midrule
Adafactor & 24 (768)  &  7.04 GiB & 39.65 $\pm$ 0.002  \\
\NAME & 24 (768)  &  7.02 GiB & $\bm{40.50 \pm 0.001}$ \\
\bottomrule
\end{tabular}
\end{small}
\end{center}
\caption{BLEU scores and memory usage for various batch sizes on the WMT'14
  \entofr dataset.}
\label{tbl:bleu}
\end{table}

\subsection{Language modeling}

Next, we considered a language modeling task on the concatenation of Wikipedia
and BooksCorpus~\cite{zhu2015aligning}, with 2.5B and 800M words respectively.
We used the recent Bidrectional Encoder Representation (BERT) architecture of
\citet{devlin18}, focusing on its larger variant, coined BERT-Large.
BERT-Large is a large bidirectional transformer model containing 24 transformer
blocks with 1024 hidden dimensions and 16 self attention heads. It has 340M
parameters (1.297 GiB), and is set up to jointly optimize two objectives: (a)
masked language model (Masked-LM) loss where the task is to predict masked
tokens based on surrounding context, and (b) next sentence prediction (NSP)
loss where the task is to predict whether two given sentences are consecutive
in the text.

\begin{figure}[ht]
\begin{center}
\includegraphics[width=0.45\columnwidth]{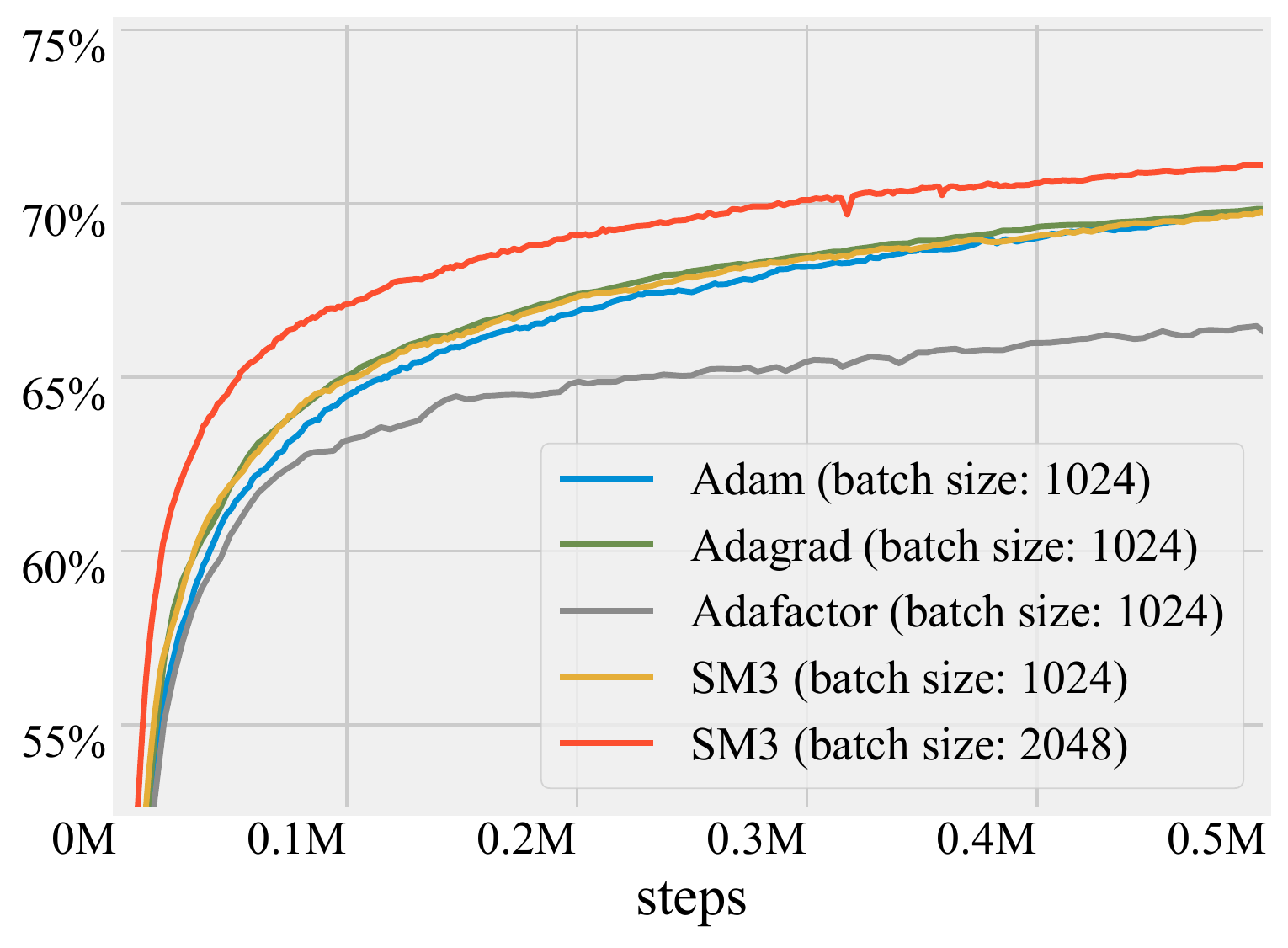}
\hspace{0.03\columnwidth}
\includegraphics[width=0.45\columnwidth]{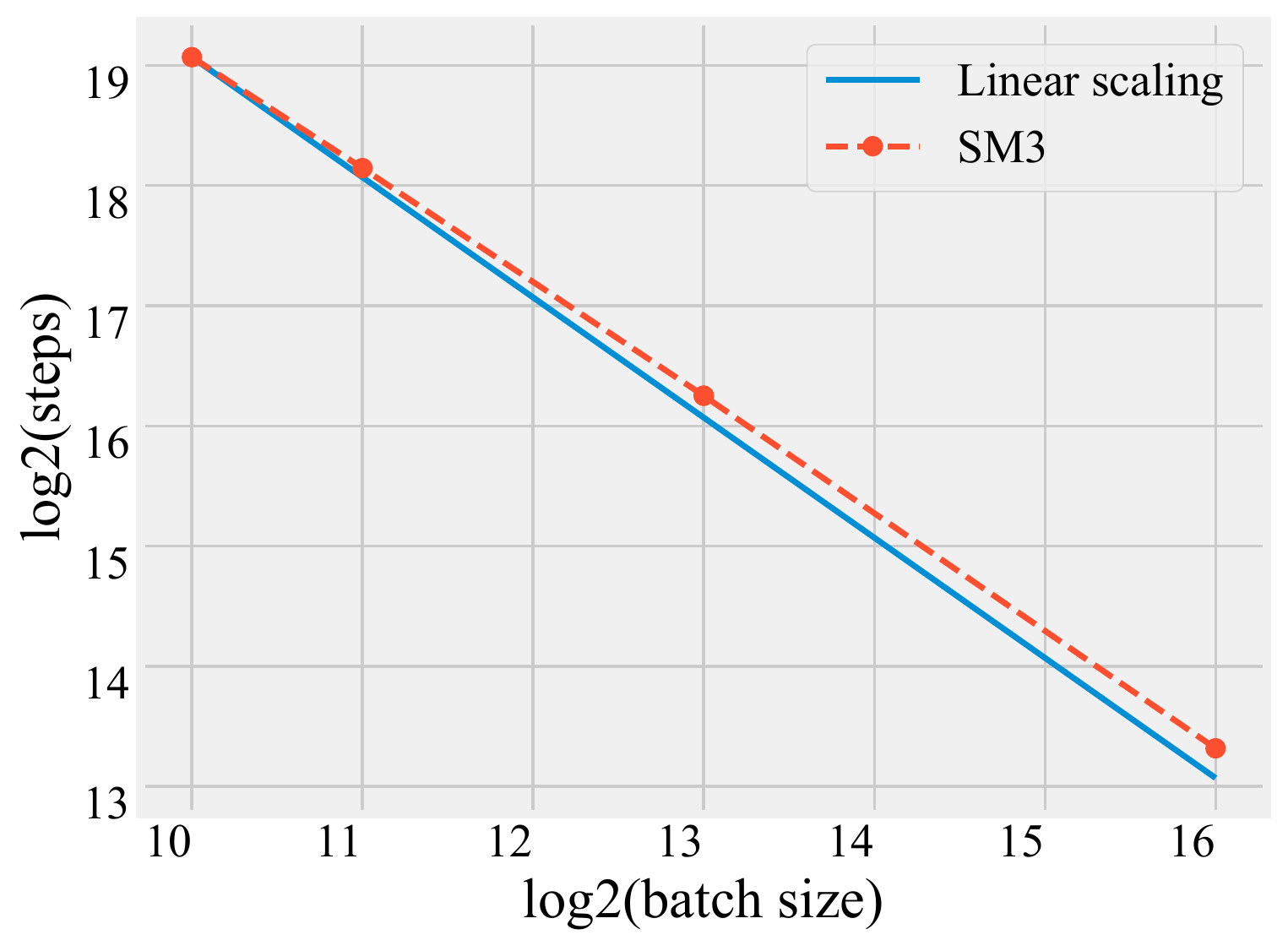}
\end{center}
\vspace{-0.5cm}
\caption{%
Masked LM test accuracy (left), and number of steps to get $70\%$ test accuracy
as a function of the batch size (right), of the BERT-Large language model
trained on Wikipedia+BooksCorpus.
\NAME with batch size 2048 uses about the same amount of memory as Adam/Adagrad
with batch size 1024, and scales linearly up to a batch size of $2^{16}$, at
which point we hit the hardware memory limits.}
\label{bert_large_loss}
\end{figure}

As before, we compared \NAME with Adagrad, Adam and Adafactor. Our results are
presented in~\cref{bert_large_loss}. We see that \NAME worked as well as Adam
and Adagrad for a fixed batch size. However, the savings in memory
allowed us to train \NAME with double the batch size, resulting in a substantial
increase in accuracy. The experiments were run using the open sourced code from
\cite{devlin18} on a $\8x8$ Cloud TPU-V2 configuration.

To underscore the importance of our memory savings in the context of very large
models, we report additional results on the number of steps required for
reaching a given solution quality for various batch sizes. We chose a solution
quality of $70\%$ Masked-LM accuracy on the holdout set, which Adam and AdaGrad
reached at 500k steps. We use Cloud TPU-v3 device which has 16Gib per core for
this experiment. We measured the number of steps \NAME needed to reach this
accuracy as a function of the batch size. Our results are presented in
\cref{bert_large_loss}. \NAME scaled almost linearly with the batch size, up to
a size of $2^{16}$, at which point the training program reached the limits of
memory available on hardware. We also found that \NAME came out ahead in terms
of wall-time: with the same batch size, a step of \NAME was faster than Adam's
by~3\%, and doubling the batch size allowed it to reach the same solution
quality in almost 35\% less wall-time for the same computational budget.

\begin{table}[t]
\vspace{-0.2cm}
\begin{center}
\begin{small}
\begin{tabular}{lcccr}
\toprule
\head{2cm}{Optimizer} & \head{3cm}{Batch Size \mbox{per core (total)}} &  \head{3cm}{Memory Usage per core}  \\
\midrule
Adam &  8 (1024) & 6.15 GiB \\
\NAME &  8 (1024) &  4.90 GiB \\
\midrule
\NAME & 16 (2048) & 6.02 GiB \\
\bottomrule
\end{tabular}
\end{small}
\end{center}
\caption{Training memory consumption at different batch sizes for BERT-Large on 8x8 TPUs.}
\label{tbl:bert}
\end{table}

\subsection{AmoebaNet-D on ImageNet}
\begin{figure}[h!]
\begin{center}
\includegraphics[width=0.45\columnwidth]{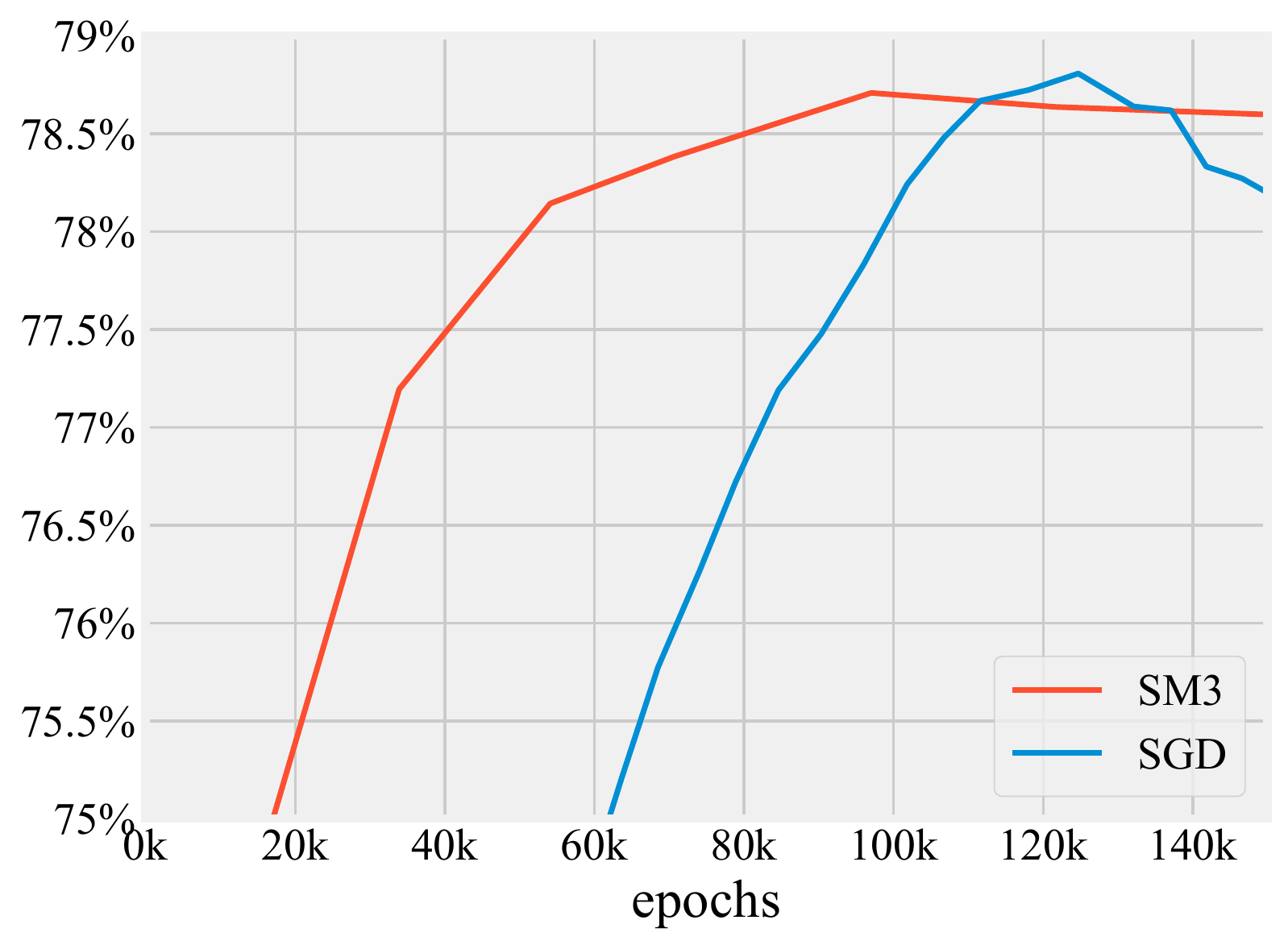}
\hspace{0.05\columnwidth}
\includegraphics[width=0.45\columnwidth]{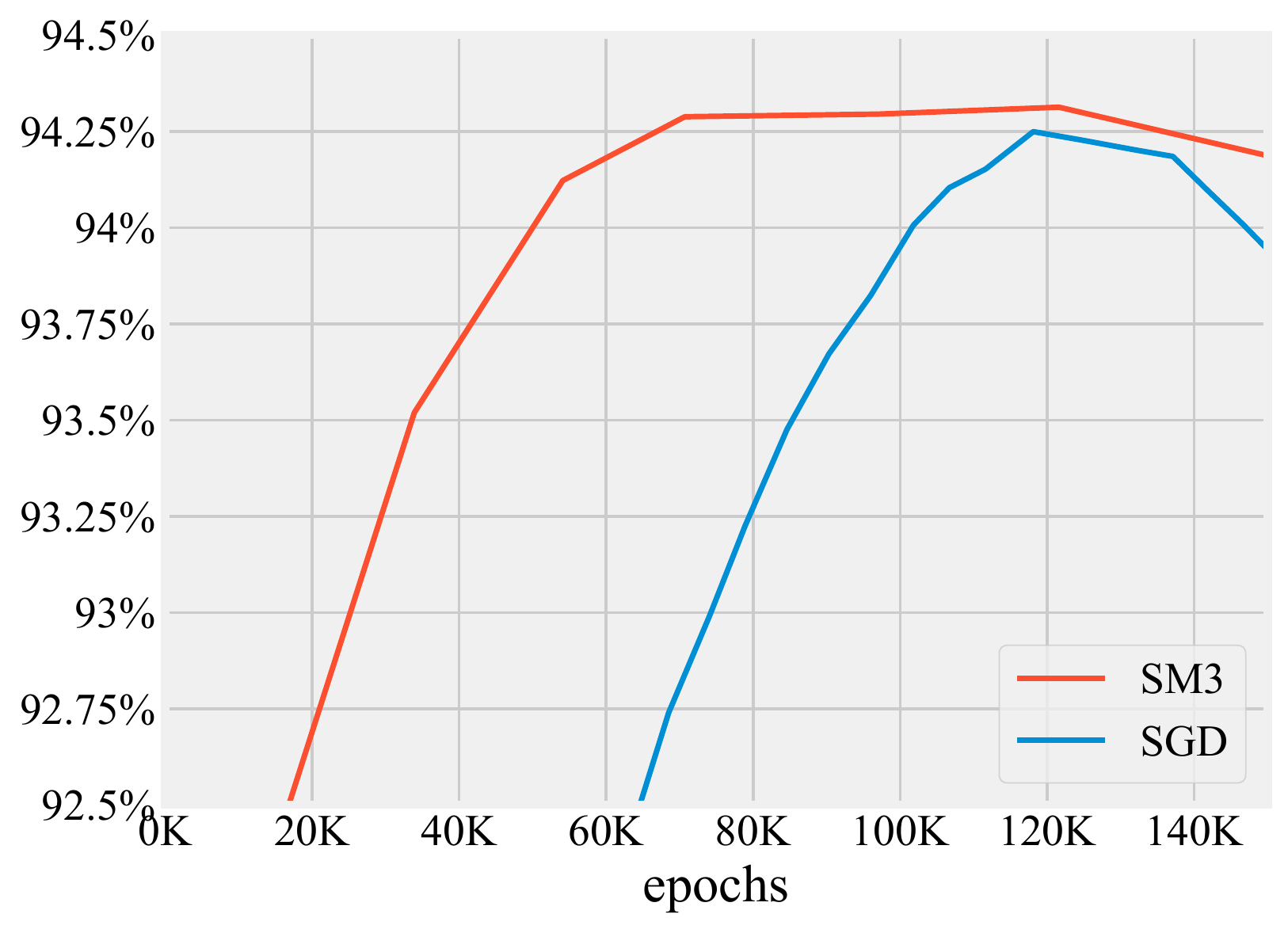}
\end{center}
\vspace{-0.5cm}
\caption{%
Top-1 (left) and Top-5 (right) test accuracy of AmoebaNet-D on ImageNet.}
\label{amoeba_net}
\end{figure}

Finally, we report results from a different domain: image classification on
ImageNet~\citep{russakovsky2015imagenet} with the state-of-the-art AmoebaNet-D
architecture~\citep{real2018regularized}, that has recently won the Stanford
DAWNBench competition~\citep{coleman2018analysis}.
We compared \NAME with SGD with momentum (Adam performed poorly on this task).
The results shown in~\cref{amoeba_net} indicate that \NAME performed very well
in this task and achieved improved convergence to state-of-the-art performance,
achieving $78.71\%$ top-1 and $94.31\%$ top-5 test accuracies.

\section{Summary}

Motivated by the large increase in models sizes and the huge amounts of memory
required for training them, we have presented a new memory-efficient adaptive
optimization algorithm for stochastic optimization called \NAME.  We
demonstrated empirically that \NAME can be used effectively in training modern
mammoth-sized models and dramatically decrease memory overhead. Utilizing the
freed memory for increasing the batch size, our experiments indicate that this
saving can also lead to significant improvements in performance.
Our theoretical investigation focused on convex objectives. As with many other
optimization scenarios, we believe the analysis of convex memory-efficient
adaptive optimization could serve as a basis for understanding non-convex
settings.

Our memory savings virtually eliminate the overhead coming
from the second-order statistics $\gamma_t$ with little and often no impact on
convergence. Additional and potentially substantial improvements in memory
consumption could come from compressing or sketching the momentum terms
employed by virtually all first-order optimizers used in practice. We leave the
exploration of this promising direction for future work.

\bibliography{lather}

\begin{thebibliography}{26}
\providecommand{\natexlab}[1]{#1}
\providecommand{\url}[1]{\texttt{#1}}
\expandafter\ifx\csname urlstyle\endcsname\relax
  \providecommand{\doi}[1]{doi: #1}\else
  \providecommand{\doi}{doi: \begingroup \urlstyle{rm}\Url}\fi

\bibitem[Abadi et~al.(2016)Abadi, Barham, Chen, Chen, Davis, Dean, Devin,
  Ghemawat, Irving, Isard, Kudlur, Levenberg, Monga, Moore, Murray, Steiner,
  Tucker, Vasudevan, Warden, Wicke, Yu, and Zheng]{tensorflow}
M.~Abadi, P.~Barham, J.~Chen, Z.~Chen, A.~Davis, J.~Dean, M.~Devin,
  S.~Ghemawat, G.~Irving, M.~Isard, M.~Kudlur, J.~Levenberg, R.~Monga,
  S.~Moore, D.~G. Murray, B.~Steiner, P.~Tucker, V.~Vasudevan, P.~Warden,
  M.~Wicke, Y.~Yu, and X.~Zheng.
\newblock Tensorflow: A system for large-scale machine learning.
\newblock In \emph{12th USENIX Symposium on Operating Systems Design and
  Implementation (OSDI 16)}, pages 265--283, 2016.

\bibitem[Agarwal et~al.(2018)Agarwal, Bullins, Chen, Hazan, Singh, Zhang, and
  Zhang]{GGT}
N.~Agarwal, B.~Bullins, X.~Chen, E.~Hazan, K.~Singh, C.~Zhang, and Y.~Zhang.
\newblock The case for full-matrix adaptive regularization.
\newblock \emph{CoRR}, abs/1806.02958, 2018.

\bibitem[Anil et~al.(2019)Anil, Gupta, Koren, and Singer]{sm3git2019}
R.~Anil, V.~Gupta, T.~Koren, and Y.~Singer.
\newblock {SM3} tensorflow optimizer.
\newblock
  \url{https://github.com/google-research/google-research/tree/master/sm3},
  2019.

\bibitem[Auer et~al.(2002)Auer, Cesa-Bianchi, and Gentile]{auer2002adaptive}
P.~Auer, N.~Cesa-Bianchi, and C.~Gentile.
\newblock Adaptive and self-confident on-line learning algorithms.
\newblock \emph{Journal of Computer and System Sciences}, 64\penalty0
  (1):\penalty0 48--75, 2002.

\bibitem[Cesa-Bianchi et~al.(2004)Cesa-Bianchi, Conconi, and
  Gentile]{cesa2004generalization}
N.~Cesa-Bianchi, A.~Conconi, and C.~Gentile.
\newblock On the generalization ability of on-line learning algorithms.
\newblock \emph{IEEE Transactions on Information Theory}, 50\penalty0
  (9):\penalty0 2050--2057, 2004.

\bibitem[Chen et~al.(2018)Chen, Firat, Bapna, Johnson, Macherey, Foster, Jones,
  Schuster, Shazeer, Parmar, Vaswani, Uszkoreit, Kaiser, Chen, Wu, and
  Hughes]{chen18}
M.~X. Chen, O.~Firat, A.~Bapna, M.~Johnson, W.~Macherey, G.~Foster, L.~Jones,
  M.~Schuster, N.~Shazeer, N.~Parmar, A.~Vaswani, J.~Uszkoreit, L.~Kaiser,
  Z.~Chen, Y.~Wu, and M.~Hughes.
\newblock The best of both worlds: Combining recent advances in neural machine
  translation.
\newblock In \emph{Proceedings of the 56th Annual Meeting of the Association
  for Computational Linguistics, {ACL} 2018}, pages 76--86, 2018.

\bibitem[Coleman et~al.(2018)Coleman, Kang, Narayanan, Nardi, Zhao, Zhang,
  Bailis, Olukotun, Re, and Zaharia]{coleman2018analysis}
C.~Coleman, D.~Kang, D.~Narayanan, L.~Nardi, T.~Zhao, J.~Zhang, P.~Bailis,
  K.~Olukotun, C.~Re, and M.~Zaharia.
\newblock Analysis of dawnbench, a time-to-accuracy machine learning
  performance benchmark.
\newblock \emph{arXiv preprint arXiv:1806.01427}, 2018.

\bibitem[Devlin et~al.(2018)Devlin, Chang, Lee, and Toutanova]{devlin18}
J.~Devlin, M.~Chang, K.~Lee, and K.~Toutanova.
\newblock {BERT:} pre-training of deep bidirectional transformers for language
  understanding.
\newblock \emph{CoRR}, abs/1810.04805, 2018.

\bibitem[Duchi et~al.(2011)Duchi, Hazan, and Singer]{duchi2011adaptive}
J.~Duchi, E.~Hazan, and Y.~Singer.
\newblock Adaptive subgradient methods for online learning and stochastic
  optimization.
\newblock \emph{Journal of Machine Learning Research}, 12\penalty0
  (Jul):\penalty0 2121--2159, 2011.

\bibitem[Gupta et~al.(2018)Gupta, Koren, and Singer]{shampoo-icml}
V.~Gupta, T.~Koren, and Y.~Singer.
\newblock Shampoo: Preconditioned stochastic tensor optimization.
\newblock In \emph{Proceedings of the 35th International Conference on Machine
  Learning}, volume~80, pages 1842--1850, 2018.

\bibitem[Hazan(2016)]{hazan2016introduction}
E.~Hazan.
\newblock Introduction to online convex optimization.
\newblock \emph{Foundations and Trends in Optimization}, 2\penalty0
  (3-4):\penalty0 157--325, 2016.

\bibitem[Jouppi et~al.(2017)Jouppi, Young, Patil, Patterson, Agrawal, Bajwa,
  Bates, Bhatia, Boden, Borchers, et~al.]{jouppi2017datacenter}
N.~P. Jouppi, C.~Young, N.~Patil, D.~Patterson, G.~Agrawal, R.~Bajwa, S.~Bates,
  S.~Bhatia, N.~Boden, A.~Borchers, et~al.
\newblock In-datacenter performance analysis of a tensor processing unit.
\newblock In \emph{Computer Architecture (ISCA), 2017 ACM/IEEE 44th Annual
  International Symposium on}, pages 1--12. IEEE, 2017.

\bibitem[Kingma and Ba(2014)]{kingma2014adam}
D.~P. Kingma and J.~Ba.
\newblock Adam: A method for stochastic optimization.
\newblock \emph{arXiv preprint arXiv:1412.6980}, 2014.

\bibitem[McMahan and Streeter(2010)]{mcmahan2010adaptive}
H.~B. McMahan and M.~Streeter.
\newblock Adaptive bound optimization for online convex optimization.
\newblock \emph{COLT 2010}, page 244, 2010.

\bibitem[Radford et~al.(2019)Radford, Wu, Child, Luan, Amodei, and
  Sutskever]{gpt2}
A.~Radford, J.~Wu, R.~Child, D.~Luan, D.~Amodei, and I.~Sutskever.
\newblock Language models are unsupervised multitask learners.
\newblock 2019.

\bibitem[Real et~al.(2018)Real, Aggarwal, Huang, and Le]{real2018regularized}
E.~Real, A.~Aggarwal, Y.~Huang, and Q.~V. Le.
\newblock Regularized evolution for image classifier architecture search.
\newblock \emph{arXiv preprint arXiv:1802.01548}, 2018.

\bibitem[Reddi et~al.(2018)Reddi, Kale, and Kumar]{reddi2018convergence}
S.~J. Reddi, S.~Kale, and S.~Kumar.
\newblock On the convergence of adam and beyond.
\newblock 2018.

\bibitem[Russakovsky et~al.(2015)Russakovsky, Deng, Su, Krause, Satheesh, Ma,
  Huang, Karpathy, Khosla, Bernstein, et~al.]{russakovsky2015imagenet}
O.~Russakovsky, J.~Deng, H.~Su, J.~Krause, S.~Satheesh, S.~Ma, Z.~Huang,
  A.~Karpathy, A.~Khosla, M.~Bernstein, et~al.
\newblock Imagenet large scale visual recognition challenge.
\newblock \emph{International Journal of Computer Vision}, 115\penalty0
  (3):\penalty0 211--252, 2015.

\bibitem[Schuster and Nakajima(2012)]{schuster12}
M.~Schuster and K.~Nakajima.
\newblock Japanese and korean voice search.
\newblock In \emph{{ICASSP}}, pages 5149--5152. {IEEE}, 2012.

\bibitem[Shalev-Shwartz(2012)]{shalev2012online}
S.~Shalev-Shwartz.
\newblock Online learning and online convex optimization.
\newblock \emph{Foundations and Trends in Machine Learning}, 4\penalty0
  (2):\penalty0 107--194, 2012.

\bibitem[Shazeer and Stern(2018)]{shazeer18}
N.~Shazeer and M.~Stern.
\newblock Adafactor: Adaptive learning rates with sublinear memory cost.
\newblock In \emph{Proceedings of the 35th International Conference on Machine
  Learning, {ICML} 2018}, pages 4603--4611, 2018.

\bibitem[Shen et~al.()Shen, Nguyen, Wu, Chen, et~al.]{lingvo}
J.~Shen, P.~Nguyen, Y.~Wu, Z.~Chen, et~al.
\newblock Lingvo.
\newblock \url{https://github.com/tensorflow/lingvo}.

\bibitem[Vaswani et~al.(2017)Vaswani, Shazeer, Parmar, Uszkoreit, Jones, Gomez,
  Kaiser, and Polosukhin]{vaswani2017attention}
A.~Vaswani, N.~Shazeer, N.~Parmar, J.~Uszkoreit, L.~Jones, A.~N. Gomez,
  {\L}.~Kaiser, and I.~Polosukhin.
\newblock Attention is all you need.
\newblock In \emph{Advances in Neural Information Processing Systems}, pages
  5998--6008, 2017.

\bibitem[Vaswani et~al.(2018)Vaswani, Bengio, Brevdo, Chollet, Gomez, Gouws,
  Jones, Kaiser, Kalchbrenner, Parmar, Sepassi, Shazeer, and
  Uszkoreit]{tensor2tensor}
A.~Vaswani, S.~Bengio, E.~Brevdo, F.~Chollet, A.~N. Gomez, S.~Gouws, L.~Jones,
  L.~Kaiser, N.~Kalchbrenner, N.~Parmar, R.~Sepassi, N.~Shazeer, and
  J.~Uszkoreit.
\newblock Tensor2tensor for neural machine translation.
\newblock \emph{CoRR}, abs/1803.07416, 2018.
\newblock URL \url{http://arxiv.org/abs/1803.07416}.

\bibitem[Wu et~al.(2016)Wu, Schuster, Chen, Le, Norouzi, Macherey, Krikun, Cao,
  Gao, Macherey, et~al.]{wu2016google}
Y.~Wu, M.~Schuster, Z.~Chen, Q.~V. Le, M.~Norouzi, W.~Macherey, M.~Krikun,
  Y.~Cao, Q.~Gao, K.~Macherey, et~al.
\newblock Google's neural machine translation system: Bridging the gap between
  human and machine translation.
\newblock \emph{arXiv preprint arXiv:1609.08144}, 2016.

\bibitem[Zhu et~al.(2015)Zhu, Kiros, Zemel, Salakhutdinov, Urtasun, Torralba,
  and Fidler]{zhu2015aligning}
Y.~Zhu, R.~Kiros, R.~Zemel, R.~Salakhutdinov, R.~Urtasun, A.~Torralba, and
  S.~Fidler.
\newblock Aligning books and movies: Towards story-like visual explanations by
  watching movies and reading books.
\newblock In \emph{Proceedings of the IEEE international conference on computer
  vision}, pages 19--27, 2015.

\end{thebibliography}

\clearpage
\appendix

\section{Omitted Proofs}
\label{sec:moreproofs}

\subsection{Proof of \cref{lem:main}}

\begin{proof}
Monotonicity is immediate as for any $r \in [k]$ the variable
$\mx{t}(r)$ is increasing in $t$ by definition. Therefore,
$
  \mi{t}(i)
  =
  \min_{r : S_r \ni i} \mx{t}(r)
$
is also increasing for all $i \in [d]$.

Next, since $g_{s}^2(i) \leq \max_{j \in S} g_{s}^2(j)$ for any
set $S$ that contains $i$, we have
$$
  g_{s}^2(i)
  \leq
  \min_{r : S_r \ni i} \max_{j \in S_r} g_{s}^2(j)
  .
$$
Hence,
\begin{align*}
  \sum_{s=1}^t g_{s}^2(i)
  \leq
  \sum_{s=1}^t \min_{r : S_r \ni i} \max_{j \in S_r} g_{s}^2(j)
  \leq
  \min_{r : S_r \ni i} \sum_{s=1}^t \max_{j \in S_r} g_{s}^2(j)
  =
  \min_{r : S_r \ni i} \mx{t}(r)
  ~.
\end{align*}
The claim now follows since $\min_{r : S_r \ni i} \mx{t}(r) = \mi{t}(i)$.
\end{proof}

\subsection{Proof of \cref{lem:main2}}

\begin{proof}
First, in order to establish monotonicity note that the algorithm maintains
$\tmx{t}(r) = \max_{j \in S_r} \tmi{t}(j)$ for $t\geq 1$ and $r \in [k]$.
Hence, for $t\geq 1$ and $i \in [d]$ we have
\begin{align*}
  \tmi{t+1}(i)
  =
  \min_{r : S_r \ni i} \max_{j \in S_r} \tmi{t}(j) + g^2_{t+1}(i)
  \geq
  \tmi{t}(i)
  ~.
\end{align*}
Let $\gamma_t(i) = \sum_{s=1}^t g^2_{s}(i)$. We next prove by induction that
$\gamma_t(i) \leq \tmi{t}(i) \leq \mi{t}(i)$ for all $t$ and $i \in [d]$.
For $t=1$ this is true as $\tmi{1}(i) = \gamma_1(i)  = g^2_{1}(i) \leq \mi{1}(i)$ for all $i$ by \cref{lem:main}.
For the induction step, assume that $\gamma_t(i) \leq \tmi{t}(i) \leq \mi{t}(i)$ for all $i$ and write
\begin{align*}
  \tmi{t+1}(i)
  &=
  \min_{r : S_r \ni i} \max_{j \in S_r} \tmi{t}(j) + g^2_{t+1}(i)
  \\
  &\geq
  \tmi{t}(i) + g^2_{t+1}(i)
  \\
  &\geq
  \gamma_t(i) + g^2_{t+1}(i)
  \\
  &=
  \gamma_{t+1}(i)
  ~ .
\end{align*}
On the other hand, we have
\begin{align*}
  \tmi{t+1}(i)
  &=
  \min_{r : S_r \ni i} \max_{j \in S_r} \tmi{t}(j) + g^2_{t+1}(i)
  \\
  &\leq
  \min_{r : S_r \ni i} \max_{j \in S_r} \mi{t}(j) + g^2_{t+1}(i)
  \\
  &\leq
  \min_{r : S_r \ni i} \max_{j \in S_r} \mi{t}(j) + \min_{r : S_r \ni i}
  \max_{j \in S_r} g^2_{t+1}(j)
  \\
  &\leq
  \min_{r : S_r \ni i} \LR{ \max_{j \in S_r} \mi{t}(j) + \max_{j \in S_r} g^2_{t+1}(j) }
  \\
  &\leq \min_{r : S_r \ni i} \sum_{s=1}^{t+1} \max_{j \in S_r} g^2_{s}(j)
  \\
  &=
  \mi{t+1}(i) ~ ,
\end{align*}
where the final inequality follows from the fact that, for all $j \in S_r$ one
get,
\begin{align*}
  \mi{t}(j)
  =
  \min_{r' : S_{r'} \ni j} \sum_{s=1}^t \max_{j' \in S_{r'}} g^2_{s}(j')
  \leq
  \sum_{s=1}^t \max_{j' \in S_r} g^2_{s}(j')
  .
  &\qedhere
\end{align*}
\end{proof}

\section{More Experiments}
\label{sec:more-exper}

\subsection{Tightness of \NAME approximation}

\begin{figure}[h!]
  \centering
  \begin{tabular}{ccc}
    \includegraphics[width=0.3\linewidth]{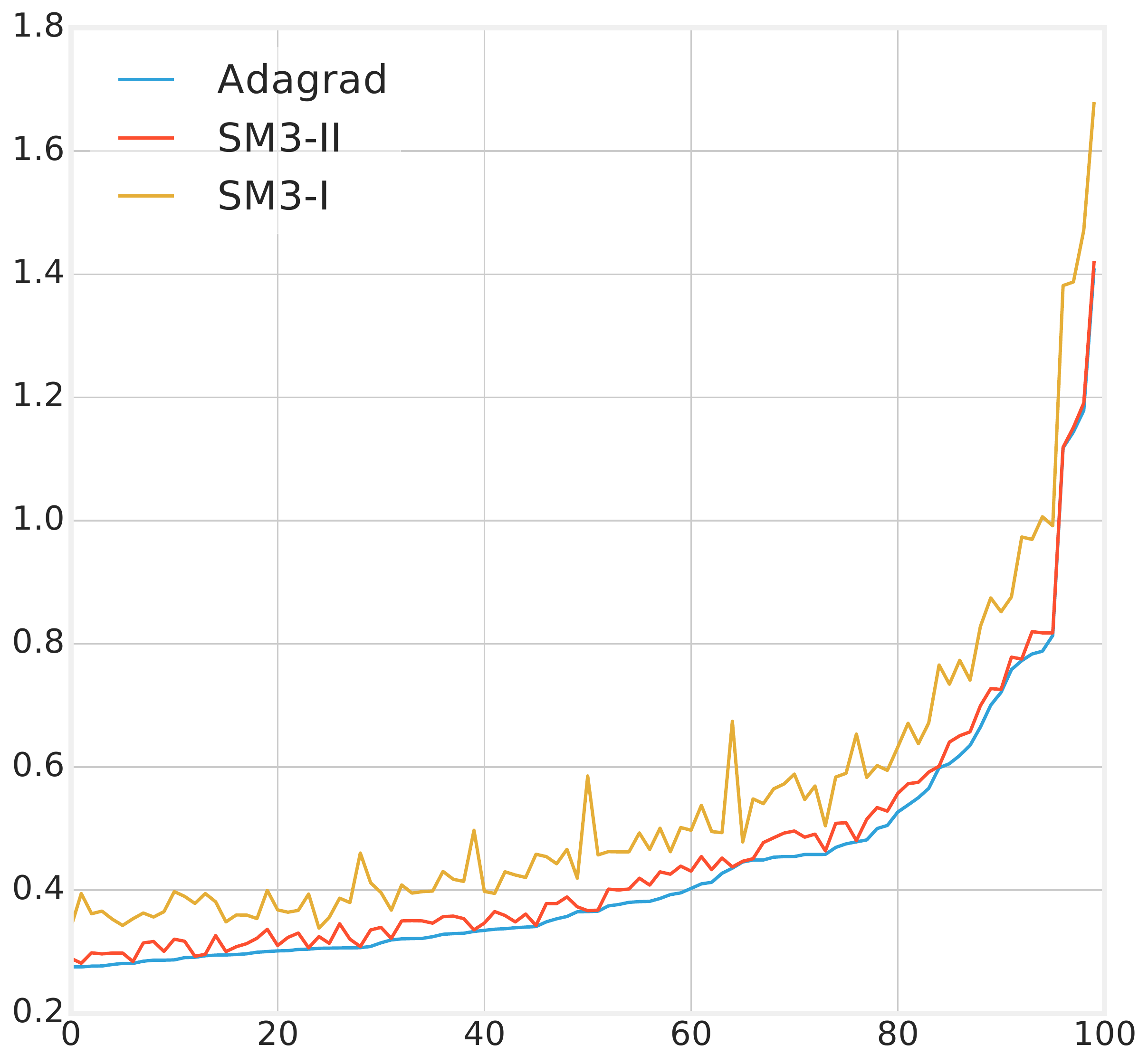} &
    \includegraphics[width=0.3\linewidth]{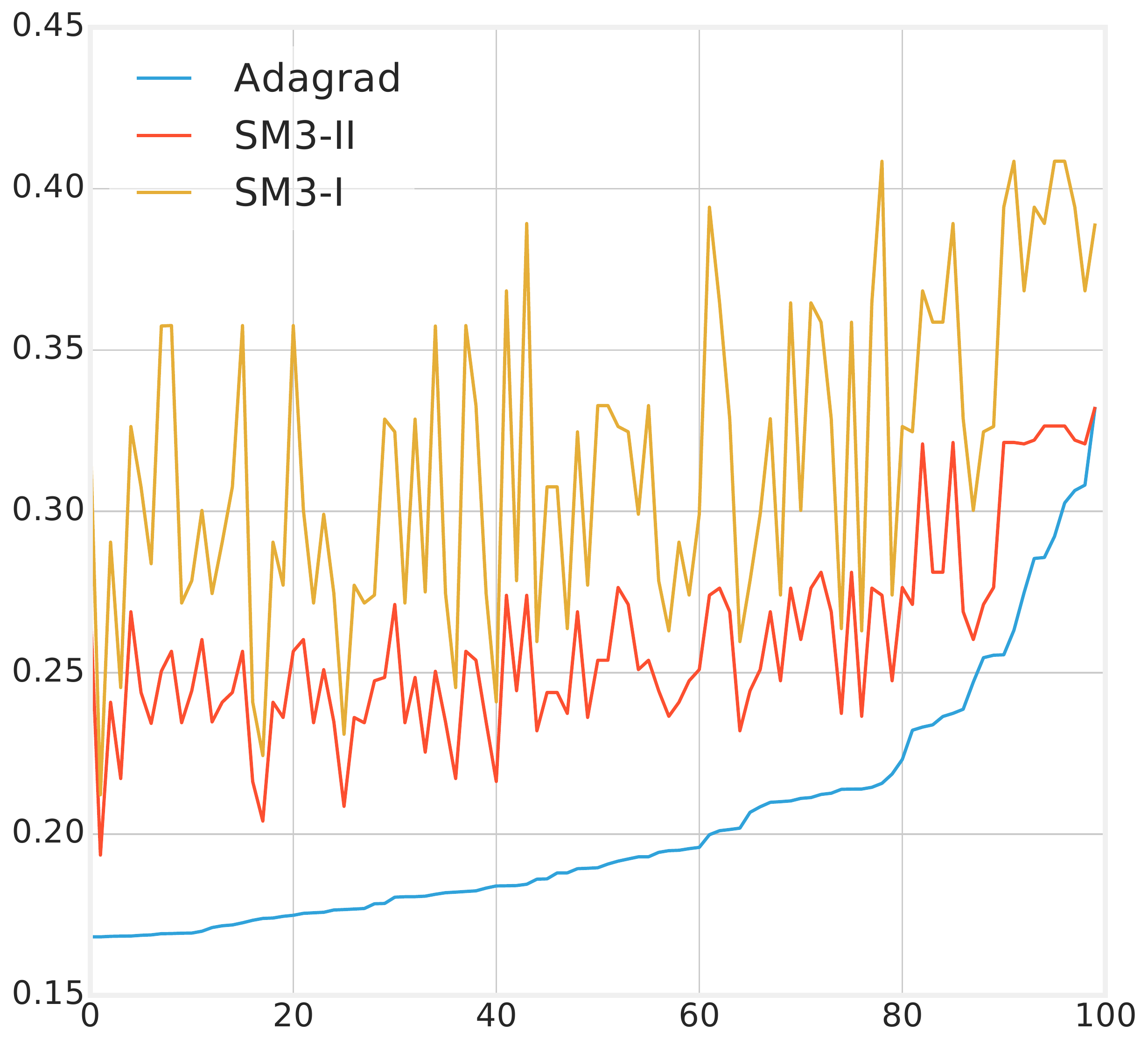} &
    \includegraphics[width=0.3\linewidth]{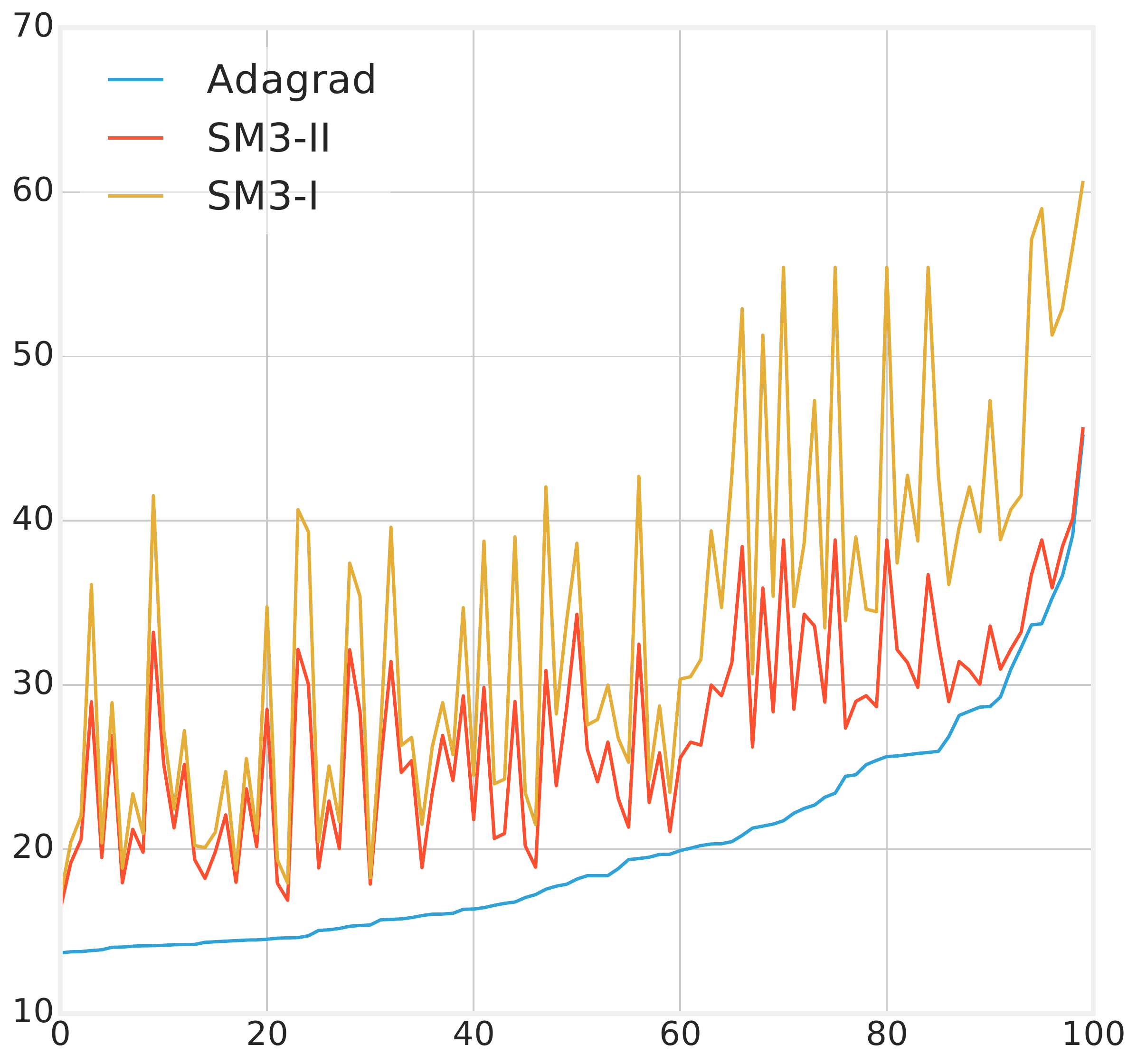}  \\
    {\small(a) Input embedding} & {\small (b) Attention layer} & {\small (c) Output softmax}
  \end{tabular}
  \caption{The magnitude of the 100 largest accumulators~\cref{eq:accum} of
	Adagrad for the embedding layer of a Transformer model trained
  on the WMT'14 \entofr dataset. The accumulators are sorted by magnitude.}
  \label{fig:sm3_2v1}
\end{figure}

We corroborate our discussion from \cref{sec:patterns} with an illustration
for both variants of \NAME, of the tightness of approximation of Adagrad's
second-order statistics. \cref{fig:sm3_2v1} demonstrates that overall \NAME
provides a tight approximation, with the \NAME-II performing significantly
better than \NAME-I, especially for higher-magnitude values.

\subsection{Results for basic Transformer on WMT'14 \entode}

\begin{figure}[h!]
\begin{center}
\begin{minipage}[c]{0.45\columnwidth}
\vspace{0pt}
\includegraphics[width=\columnwidth]{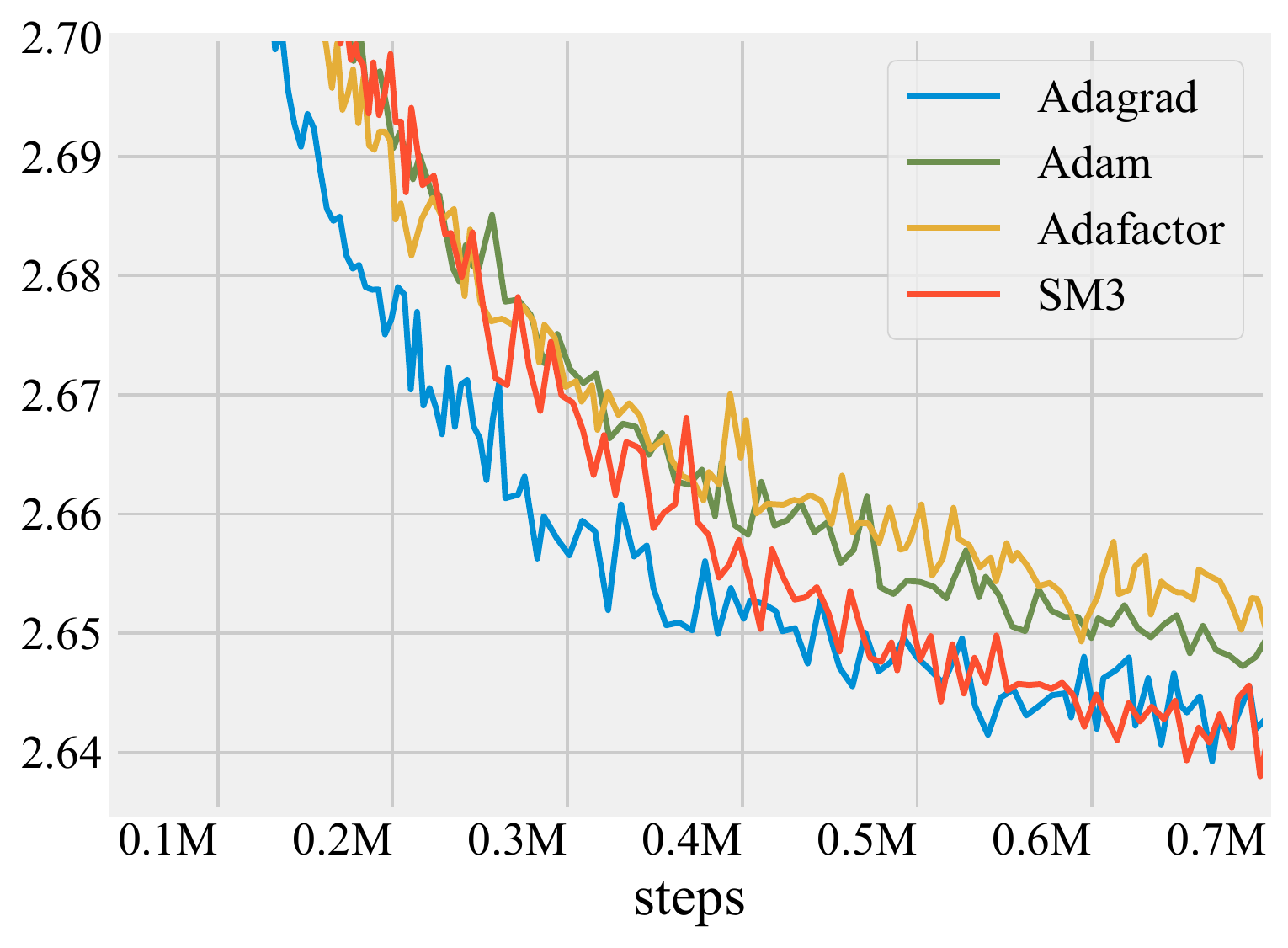}
\end{minipage}
\hspace{0.025\linewidth}
\begin{minipage}[c]{0.45\columnwidth}
\vspace{0pt}
\begin{center}
\begin{small}
\begin{tabular}{lcc}
\toprule
\head{1.25cm}{Optimizer} & \head{2.25cm}{Batch Size \mbox{p/core (total)}} &  \head{0.75cm}{BLEU} \\
\midrule
Adam & 48 (1536) & 27.15 $\pm$ 0.002 \\
Adagrad & 48 (1536) & {\bf 27.42  $\pm$ 0.001} \\
Adafactor & 48 (1536) & 26.88  $\pm$ 0.002 \\
\NAME & 48 (1536) & 27.32  $\pm$ 0.002 \\
\bottomrule
\end{tabular}
\end{small}
\end{center}
\end{minipage}
\end{center}
\caption{Test log-perplexity (left) and BLUE scores (right) on of a Transformer
model trained on the WMT'14 \entode dataset.}
\label{en_de}
\end{figure}

In~\cref{en_de} we report results for the basic Transformer after training for
700,000 steps on \entode with a batch size of 1536.  As in previously
discussed experiments, SGD with momentum performs poorly compared to adaptive
optimizers and hence is not included in the comparison.

\subsection{Activation patterns in convolutional networks}
\label{sec:conv-patterns}

\begin{figure}[ht]
  \vspace{-0.2cm}
  \centering
  \begin{tabular}{c c}
    \includegraphics[height=115pt,width=115pt,trim=0 0 0 0]{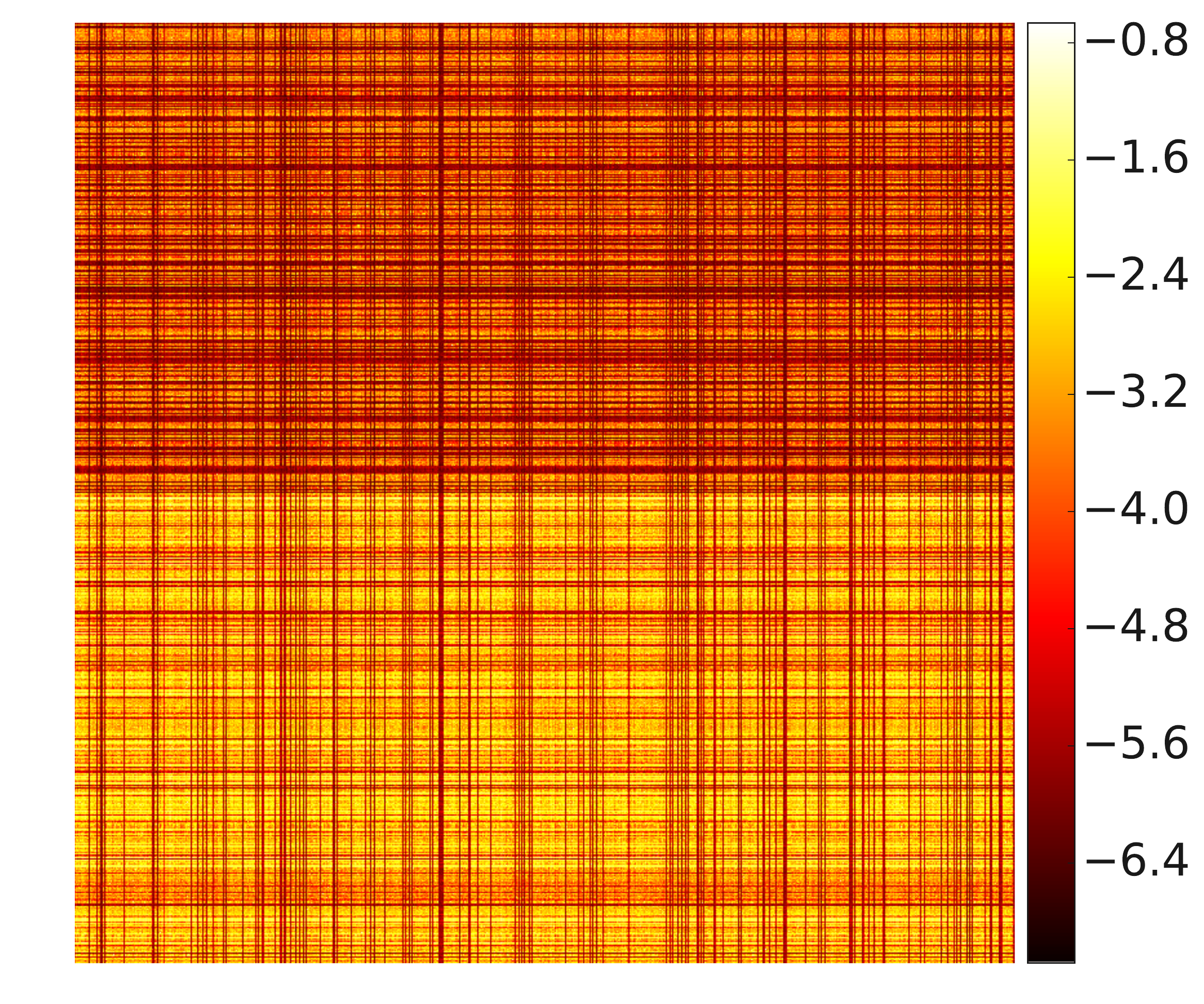} &
    \includegraphics[height=115pt,width=115pt,trim=0 0 0 0]{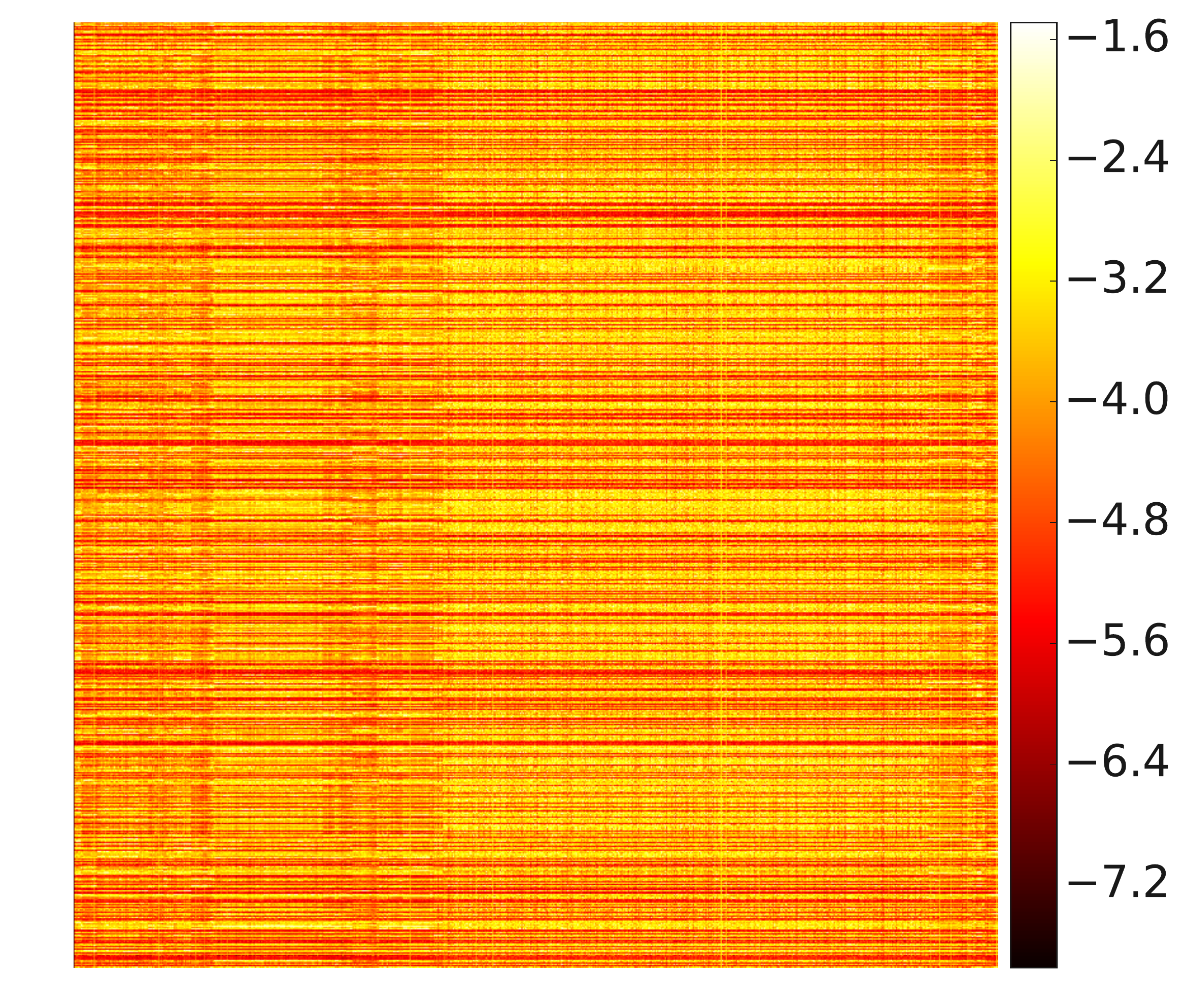} \\
    {\small (a) Filter of 7x1x256x256 convolution \hspace{15pt}} & {\small (b) Output softmax \hspace{15pt}}
  \end{tabular}
  \caption{Visualization of gradient square statistics~\cref{eq:accum} for
  different weight matrices using a AmoebaNet-D model. (Color intensities are in
  log~scale.)}
  \label{fig:adagrad_lr_conv}
\end{figure}

We give additional evidence of self-formation of row and column activation
patterns which arise in convolutional image recognition models. See
\cref{fig:adagrad_lr_conv} for an illustration.

\section{Details of Experiments}
\label{sec:exper-detail}

We report the settings of hyperparameters used in our experiments in
\cref{tbl:hparams}. We performed a grid search on the following
hyper-parameters: $\eta \in [10^{-5},10^{0}]$,  $\beta_1 \in \{0.9, 0.95,
0.99\}$, and $\beta_2 \in [0.9, 0.999]$ for each of the optimizers when
applicable. We were able to discard a large fraction of the search space for
learning rates, as large values typically cause instability and lower values
make the progress slow. We found we found $\beta_1 = 0.9$ to work well for
almost all experiments (where batch size < 2048) except in the case of \NAME
on BERT-Large where $\beta_1 = 0.95$ worked best for $2^{13}$ and $2^{16}$
batch sizes.

\begin{table}[ht]
\centering
\begin{small}
\begin{tabular}{llclc}
\toprule
\head{1.5cm}{Experiment} & \head{1cm}{Optimizer}& \head{1.5cm}{Batch size} & \head{2cm}{Hyperparameters} & \head{1.5cm}{Warmup ($T_{0}$)}    \\
\midrule
\multirow{4}{*}{Transformer \entode}  &  Adafactor & 1536 & $\eta=0.0003$, $\beta_{1}=0.9$, $\beta_{2}=0.98$ & 10k  \\
 &  Adam & 1536 & $\eta=0.0004$,  $\beta_{1}=0.9$, $\beta_{2}=0.98$ & 10k \\
&  Adagrad & 1536 & $\eta=0.1$, $\beta_{1}=0.9$  & 10k \\
&  SM3 & 1536 & $\eta=0.225$, $\beta_{1}=0.9$ & 10k \\
\midrule
\multirow{4}{*}{Transformer \entofr}  &  Adafactor& 384 & $\eta=0.00045$, $\beta_{1}=0.9$, $\beta_{2}=0.98$ & 40k    \\
 &  Adam& 384 & $\eta=0.00015$, $\beta_{1}=0.9$, $\beta_{2}=0.98$ & 40k \\
&  Adagrad& 384 & $\eta=0.075$, $\beta_{1}=0.9$  & 40k\\
&  SM3& 384 & $\eta=0.125$, $\beta_{1}=0.9$ & 40k\\
&  Adafactor & 768 & $\eta=0.00045$, $\beta_{1}=0.9$, $\beta_{2}=0.98$  & 40k\\
&  SM3  & 768 & $\eta=0.25$, $\beta_{1}=0.9$ & 40k\\
\midrule
\multirow{4}{*}{BERT--Large}  &  Adafactor  & 1024 & $\eta=0.005$, $\beta_1=0.9$, $\beta_2=0.999$ & 10k  \\
 &  Adam  & 1024 & $\eta=0.0001$, $\beta_1=0.9$, $\beta_2=0.999$ & 10k\\
&  Adagrad  & 1024 & $\eta=0.25$, $\beta_1=0.9$ & 10k \\
&  SM3  & 1024 & $\eta=0.1$, $\beta_1=0.9$   & 10k\\
&  SM3  & 2048 & $\eta=0.1$, $\beta_1=0.9$  & 10k\\
&  SM3  & 8192 &  $\eta=0.05$, $\beta_1=0.95$ & 2k\\
&  SM3 & 65536 &  $\eta=0.15$, $\beta_1=0.95$ & 2k \\
\midrule
\multirow{2}{*}{AmoebaNet}  &  SGD  & 4096& $\eta=6.15$, $\eta_0=0.042$, $\tau$=4.5k, $\beta_1=0.9$ & 1.2k\\
					   &  SM3 & 4096 & $\eta=0.5$, $\beta_1=0.9$ & 1.2k\\
\bottomrule
\end{tabular}
\end{small}
\caption{Hyperparameter setup used in our experiments.}
\label{tbl:hparams}
\end{table}

We also employed a short initial ``warmup'' stage for all optimizers. During
warmup we gradually increased the learning rate $\eta$ from zero to its
maximal value during the first few thousand updates. This is a common
heuristic in training of deep models, where often a high learning rate setting
in the early stages of optimization causes instabilities and results in
failure to converge, colloquially called ``blowup''.
The choice of the number of steps used for warmup does not affect the eventual
performance of the trained models, and was chosen somewhat liberally before
tuning the rest of the hyperparameters. We would like to stress that in each
experiment, we used the same value for all optimizers. In the experiments with
BERT-Large using large batch sizes, warmup was very short as experimentally it
deemed almost unnecessary.

\begin{table}[ht]
\begin{center}
\begin{small}
\begin{tabular}{lllc}
\toprule
\head{2.5cm}{Experiment} & \head{2.5cm}{Optimizer} & \head{3cm}{LR Schedule (after warmup)} & \head{2cm}{Reference}\\
\midrule
Transformer & Adam, Adafactor & $\eta \sqrt{d/t}$ & \cite{vaswani2017attention} \\
BERT & Adam, Adafactor & $\eta(1 - t/T)$ & \cite{devlin18} \\
AmoebaNet-D & SGD (+momentum) & $\max\Lrset{\eta_0, \eta \alpha^{\floor{t/\tau\!}}}$ & Folklore \\
All & Adagrad, SM3 & $\eta$ & \\
\bottomrule
\end{tabular}
\end{small}
\end{center}
\caption{Learning rate schedules used by the algorithms we experimented with.
Here, $t$ is the current time step, $\eta$ is the base learning rate, $\alpha<1$
is the decay constant, $\tau$ is the staircase step interval, $\eta_0$ is the
minimum learning rate for staircase schedule, $T_{0}$ is the number of warmup
steps, $T$ is the total number of training steps, and $d$ is the size of the model.}
\label{learning_rate_rules}
\end{table}

We note that, compared to other optimizers, \NAME has a \emph{single}
hyper-parameter that requires tuning, the initial learning rate $\eta$.
Concretely, past the warmup phase, \NAME does \emph{not} employ a schedule for
learning-rate decay which is often difficult to tune. \cref{learning_rate_rules}
we summarize the procedures for scheduling the learning rate of all
optimizers.

For experimenting with Adafactor, we made use of the implementation in the
Tensor2Tensor framework~\cite{tensor2tensor} and tuned the parameters as
described above. We found Adafactor to work quite well on translation tasks,
for which it was designed and optimized. Alas, we could not get it to work on
the BERT language models. Adafactor's implementation has numerous
hyperparameters which makes it extremely difficult to set up for new domains.

\end{document}